\documentclass[10pt,twocolumn,letterpaper]{article}

\usepackage{iccv}
\usepackage{times}
\usepackage{epsfig}
\usepackage{graphicx}
\usepackage{amsmath}
\usepackage{amssymb}

\usepackage[pagebackref=true,breaklinks=true,letterpaper=true,colorlinks,bookmarks=false]{hyperref}

\usepackage{graphicx}
\usepackage{amsmath}
\usepackage{amssymb}
\usepackage{booktabs}

\usepackage{algorithm,algpseudocode}
\usepackage[algo2e]{algorithm2e} 
\usepackage{multirow}
\usepackage{threeparttable}
\usepackage{diagbox}
\usepackage[table]{xcolor}
\usepackage{caption}
\usepackage{subcaption}
\usepackage{authblk}

\iccvfinalcopy 

\def\iccvPaperID{xxxx} 

\ificcvfinal\fi

\title{AugDiff: Diffusion based Feature Augmentation for Multiple Instance Learning in Whole Slide Image\vspace{-1em}}

\author{%
\small
  Zhuchen Shao\textsuperscript{1}, Liuxi Dai\textsuperscript{2}, Yifeng Wang\textsuperscript{2}, Haoqian Wang\textsuperscript{1 ${\ast}$}, Yongbing Zhang\textsuperscript{2}\thanks{Corresponding authors.}\\
  \vspace{-0.5em}
  \textsuperscript{1}Tsinghua Shenzhen International Graduate School, Tsinghua University\\ \textsuperscript{2}Harbin Institute of Technology (Shenzhen)\\
  \texttt{shaozc0412@gmail.com}, \texttt{dailiuxi21@stu.hit.edu.cn},
  \texttt{wangyifeng@stu.hit.edu.cn},
  \texttt{wanghaoqian@tsinghua.edu.cn}, 
  \texttt{ybzhang08@hit.edu.cn} \vspace{-2.5em}\\
}

\begin{document}

\maketitle
\ificcvfinal\thispagestyle{empty}\fi

\begin{abstract}
Multiple Instance Learning (MIL), a powerful strategy for weakly supervised learning, is able to perform various prediction tasks on gigapixel Whole Slide Images (WSIs). 
However, the tens of thousands of patches in WSIs usually incur a vast computational burden for image augmentation, limiting the MIL model's improvement in performance. Currently, the feature augmentation-based MIL framework is a promising solution, while existing methods such as Mixup often produce unrealistic features.  
To explore a more efficient and practical augmentation method, we introduce the Diffusion Model (DM) into MIL for the first time and propose a feature augmentation framework called AugDiff. 
Specifically, we employ the generation diversity of DM to improve the quality of feature augmentation and the step-by-step generation property to control the retention of semantic information.
We conduct extensive experiments over three distinct cancer datasets, two different feature extractors, and three prevalent MIL algorithms to evaluate the performance of AugDiff. 
Ablation study and visualization further verify the effectiveness.
Moreover, we highlight AugDiff's higher-quality augmented feature over image augmentation and its superiority over self-supervised learning. The generalization over external datasets indicates its broader applications. 

\end{abstract}

\begin{figure}[h]
    \centering
    \includegraphics[width=0.95\linewidth]{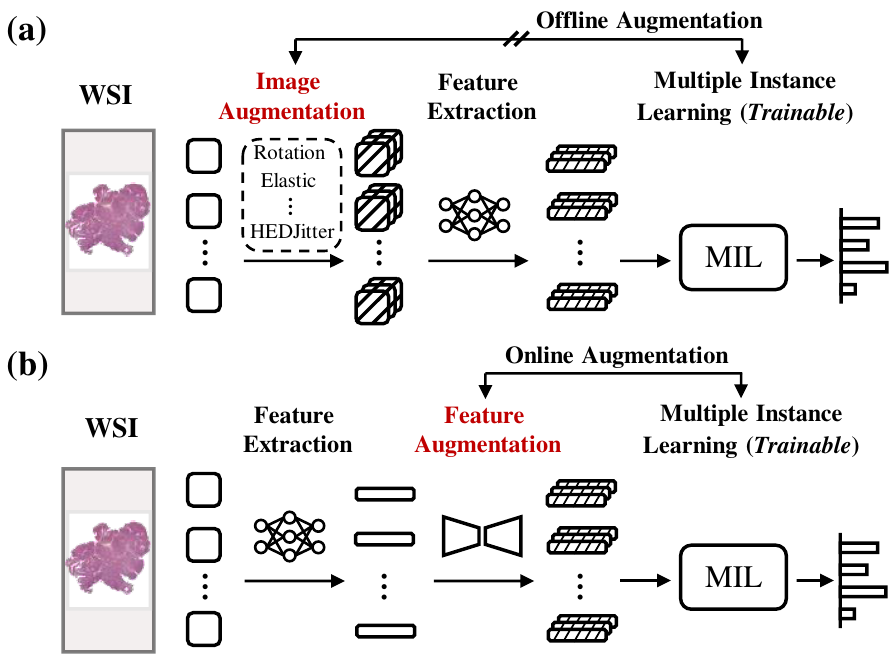}
    \caption{\textbf{Background and Motivation.} 
        \textbf{(a)} 
The standard framework for combining MIL and WSI-related tasks:
1)~The gigapixel WSI is sliced into several patches. 2)~The patches are used for image augmentation before being embedded into features. 3)~The training of the MIL model is supervised by WSI-level labels.
It should be noted that image augmentation can only be utilized offline to assist with MIL.
        \textbf{(b)} The framework facilitates MIL training through feature augmentation. In contrast to (a), feature augmentation is integrated into MIL training online, preventing repeating feature extraction while allowing for real-time augmentation.
        Note that quantitative changes in the figure depict the expansion of the dataset. The training dataset for (a) is expanded in the image augmentation,  whereas the training dataset for (b) is expanded in the feature augmentation.
    }
    \label{fig:feature_augmentation}
\end{figure}
\section{Introduction}
	Computational pathology is a research direction that combines deep learning with high-resolution, wide-field-of-view Whole Slide Images (WSIs) \cite{srinidhi2021deep, komura2018machine}. 
 An important thing to figure out is how to design a deep learning network for acquiring information from gigapixel images. Based only on WSI-level annotation, Multiple Instance Learning (MIL) is currently a widely used solution paradigm \cite{lu2021ai, lu2021data, campanella2019clinical, shao2021transmil, chen2021whole, yao2020whole}.
	
As a method of weakly supervised learning, MIL needs to learn the relationship from many unlabeled instances to bag-level labels. Similarly, WSI-related tasks require aggregating numerous unlabeled patches to predict WSI-level labels. Since the tens of thousands of patches in WSIs can impose a tremendous computing burden, MIL is frequently based on aggregating patch features. Unfortunately, limited by the small number of WSI-level training data, the performance of MIL methods is always sub-optimal \cite{shao2021weakly, zhang2022dtfd}. 
Intuitively, image augmentation can efficiently overcome this issue, while image augmentation and feature extraction over tens of thousands of patches are time-consuming and computationally demanding. As shown in Figure \ref{fig:feature_augmentation}(a), the image augmentation can only facilitate MIL training offline, which cannot effectively improve feature diversity. 
Additionally, traditional image augmentation methods often make limited adjustments at the feature level due to the vast amount of redundant and irrelevant information in digital images \cite{Rombach_2022_CVPR, he2022masked}.

Some research has begun to concentrate on the framework of feature augmentation. As shown in Figure \ref{fig:feature_augmentation}(b), the feature augmentation framework avoids repeated feature extraction. The extracted patch features can be repeatedly augmented in each epoch of MIL training via online augmentation. Some existing methods include the Mixup-based feature mixing framework \cite{gadermayr2022mixup, yang2022remix} and the GAN-based feature generation framework \cite{zaffar2022embedding}. However, limited by unrealistic generation and unstable performance, current feature augmentation frameworks do not always perform better than the image augmentation framework. As a result, how to achieve the goal of faster speed and better performance within the MIL feature augmentation framework remains an urgent problem to be solved.

Currently, the Diffusion Model (DM) \cite{ho2020denoising} is an innovative generative modeling solution. It outperformed GAN in various image generation tasks, and its generation is more diverse \cite{dhariwal2021diffusion, ozbey2022unsupervised, song2020score, croitoru2022diffusion}. 
 Furthermore, the Latent Diffusion Model \cite{Rombach_2022_CVPR} shows that DM could also generate various features and achieve satisfactory results in downstream tasks. Inspired by this, we consider introducing the diversity-generating ability of DM into MIL.
 It should be noted that the feature augmentation should avoid destroying the basic semantic information, \eg, the features of the normal patch ought not to be transformed into those of the tumor patch. Hence, how to flexibly adjust Diffusion for semantic information retention and feature augmentation is a crucial consideration in the framework design.

We propose a Diffusion-based feature augmentation framework, AugDiff. The main contributions of our work include the following:

1) We use the DM in combination with MIL training for the first time. Specifically, we integrate the diversity-generating capability of the DM with the online augmentation capability of the feature augmentation, which can efficiently and effectively improve MIL training.

2) We guide the training of DM with various image augmentations to assist DM in generating diverse feature augmentation. We exploit the step-by-step generation characteristic of DM to control the retention of semantic information during the augmentation sampling process.

3) We conduct extensive experiments on three distinct cancer datasets, two different feature extractors, and three prevalent MIL algorithms. Our AugDiff achieves the highest AUC metric performance compared to existing MIL augmentation frameworks. In addition, we demonstrat the significant speed advantage of AugDiff over offline Patch Augmentation. Ablation study and visualization further verify the effectiveness of AugDiff.
 
 4) To demonstrate the superiority of AugDiff, we illustrate the higher quality of AugDiff-augmented features than Patch Augmentation. Moreover, We highlight the superiority of image augmentation-guided AugDiff over the agent task-guided self-supervised learning framework. Also, cross-tests with different cancer datasets show that AugDiff has a solid ability to generalize.

\section{Related Work}

	Deep learning typically requires a large amount of labeled training data to prevent network over-fitting \cite{shorten2019survey}. Unfortunately, in many fields, including medical image analysis, large quantities of annotated data are not always available \cite{chlap2021review}. At present, data augmentation is a low-cost method for enhancing the quantity and quality of training data sets, which is widely applied in the field of medical image analysis \cite{van2021deep, kleppe2021designing, chen2022generative}.
	
	Existing research on data augmentation mainly includes non-parametric augmentation (such as rotation, stretching, etc.) and parametric augmentation (such as GAN and VAE). For the non-parametric augmentation, some studies \cite{tellez2019quantifying, abdollahi2020data} have discussed the influence of common image augmentation, such as color augmentation and normalization, on multiple WSI-related downstream tasks. 
	For the parametric image augmentation, some studies \cite{xue2021selective, chen2021synthetic} have discussed the efficacy of the GAN-based image synthesis framework for downstream classification tasks.

	As a typical weakly supervised learning method, the performance of MIL is usually constrained by the limited labeled datasets in WSI-related tasks. Due to the properties of gigapixels in WSI, traditional image-based data augmentation methods are typically computationally intensive and time-consuming. Consequently, some studies have investigated the combination of feature augmentation and MIL, including methods based on patch feature mixing and patch feature generation. 	
 The feature mixing methods include feature mixup \cite{gadermayr2022mixup, yang2022remix} and feature pseudo-bag construction \cite{shao2021weakly,zhang2022dtfd}. Due to a lack of learnable parameters, this type of feature augmentation method frequently results in a lack of variation in feature synthesis. Besides, the deep learning-based feature generation schemes include GAN-based feature augmentation \cite{zaffar2022embedding}. However, an unstable feature augmentation framework can easily destroy the semantic information of the original patch features, leading to deteriorated performance of the downstream tasks. Currently, the performance of most feature augmentation methods is inferior to that of image augmentation. How to design an efficient and effective feature augmentation method that achieves better performance and higher speed than image augmentation is an important research topic.

\begin{figure*}[h]
	\centering
	\includegraphics[width=0.95\linewidth]{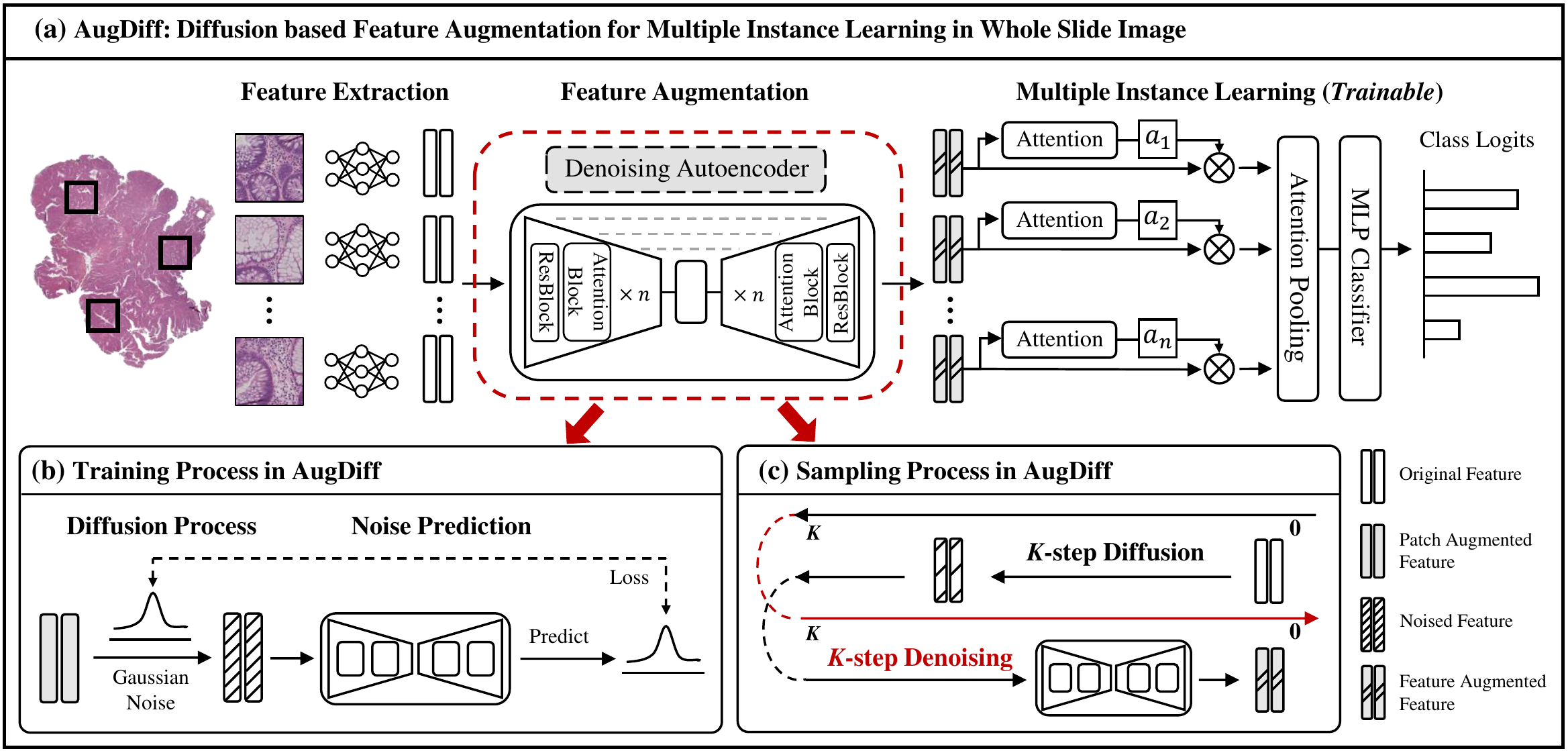}
	\caption{\textbf{Overview of AugDiff.} 
		\textbf{(a)} For tasks related to WSI, the feature-based MIL framework has two steps: pre-processing and MIL model training. Pre-processing includes WSI splitting and feature extraction. Then, during MIL training, the model learns to predict bag-level labels based on all the extracted features, and this paper adds the step of feature augmentation.
		\textbf{(b)}  The training process of the Denoising AutoEncoder (DAE) in AugDiff comprises both adding noise in the diffusion process and predicting noise by DAE.
		\textbf{(c)}  
		The sampling process in the AugDiff is a two-stage feature augmentation, which includes $K$-step Diffusion and $K$-step Denoising.
	}
	\label{fig:augdiff}
\end{figure*}


\section{Method}
\subsection{Problem Formulation}

WSI is a high-resolution, wide-field-of-view image and often only has a WSI-level label. For such weakly supervised learning problems, MIL is an efficient solution.
Here, each WSI is viewed as a bag, and each small patch cut from the WSI is considered an instance. It should be noted that the bag-level label is known, but the instance-level label is unknown.
The performance of the MIL model in WSI-related tasks is frequently constrained by a small number of training data. Data augmentation is considered a promising method to get rid of this problem.
However, every patch-based data augmentation requires unavoidable and repeated feature extraction, resulting in high computational and time costs. In addition, due to the high cost of patch augmentation, only a limited number of augmentations can be allowed, thereby limiting the variety of augmentations.
Here, feature-level augmentation is used to facilitate MIL training. The whole framework is shown in Figure \ref{fig:augdiff}.
	
\subsection{Diffusion Model based Feature Augmentation Framework in MIL Training}

	We propose AugDiff, a Diffusion-based feature augmentation framework in MIL training. This subsection will briefly introduce the Diffusion Model (DM) before focusing on the training and testing in AugDiff.

 \paragraph{Diffusion Model.}
Deep neural networks can approximate the probability distribution $p(x)$ of the data. Specifically, the DM approximates the $p(x)$ by learning the reverse process of a fixed Markov Chain of  length $T$. The forward diffusion process can be defined as successively adding Gaussian noise $\epsilon$ to the input $x$ to produce a set of noisy samples $\{x_t\}^T_{t=1}$. The reverse process can be simplified by training a Denoising AutoEncoder (DAE), $\epsilon_\theta\left(x_t, t\right)$, to predict a denoised variant of its input $x_t$ ($t = 1, \ldots, T$). The corresponding goal is as follows:
	\begin{equation}
	L_{D M}=\mathbb{E}_{x, \epsilon \sim \mathcal{N}(0,1), t}\left[\left\|\epsilon-\epsilon_\theta\left(x_t, t\right)\right\|_2^2\right].
	\end{equation}
	
	Furthermore, to be more computationally efficient, the Latent Diffusion Model (LDM) investigates training a DAE $\epsilon_\theta\left(z_t, t\right)$ at the feature level to approximate the probability distribution $p(z)$ of features.
	A conditional mechanism is also introduced into DAE to model the conditional distribution $p(z|y)$, where the conditional DAE can be denoted as $\epsilon_\theta\left(z_t, t, y\right)$. It influences the generation process by controlling the input condition $y$, which is processed using word vector mapping $\tau_\theta$ from the category to the embedding vector. The corresponding goal is as follows:
	\begin{equation}
	L_{L D M}=\mathbb{E}_{z, y, \epsilon \sim \mathcal{N}(0,1), t}\left[\left\|\epsilon-\epsilon_\theta\left(z_t, t, \tau_\theta(y)\right)\right\|_2^2\right]. \label{con:ldm}
	\end{equation}
	
MIL training in WSI-based tasks is typically performed at the feature level due to the high computational cost. Therefore, we use LDM to design a feature augmentation framework in MIL tasks.

\paragraph{Feature Augmentation based MIL Training}

We further demonstrate how DM can be utilized in MIL training. Specifically, 1) the diversity of the samples generated by the DM can assist MIL training in efficiently expanding the training data \cite{dhariwal2021diffusion}. 2) The step-by-step generation framework of DM can have better controllability in feature augmentation \cite{liew2022magicmix}. We can determine how much semantic information is kept in MIL training with different feature extractors and datasets by changing the number of feature augmentation steps. Inspired by these advantages, the proposed framework is illustrated in Algorithm \ref{alg:MIL}. The input features are first augmented by AugDiff and then provided to the MIL model to predict bag-level labels in each epoch of MIL training.  We select three classic MIL methods, including the attention-based AMIL \cite{lu2021data}, the loss-guided attention-based LossAttn \cite{shi2020loss}, and the dual-stream attention based DSMIL \cite{li2020dual}. 
It should be noted that during the MIL training process, the trained DAE in AugDiff does not participate in the optimization process.

\begin{algorithm}[ht]
	\small
	\caption{AugDiff for MIL Training}
	\label{alg:MIL}
	\KwIn{
		The bags in the training dataset $\mathbf{Z_O}=\{Z^i_O\}_{i=1}^{M}$. Bag label $\{Y^i\}_{i=1}^{M}$.
		Proposed Diffusion-Denoising sampling process $\mathcal{T}$. Trained DAE $\epsilon_\theta$. Diffusion and Denoising step $K$ in the sampling process.
	}
	\KwOut{Trained MIL model $\rho$.}
	\While{not converged}{
		\For{${Z}_O^{i}$ in  $\mathbf{Z_O}$}{
			{\%1. AugDiff performs feature augmentation on the input $Z_O^i$. Then, the MIL model takes the augmented feature $\widetilde{Z}_O^i$ as input to predict the score vector.}\\
			$p \leftarrow \rho\left(\mathcal{T}\left({Z_O^i}, \epsilon_\theta, K\right)\right)$ \\
			{\%2. Take the gradient descent step to optimize the MIL model.}\\
			$
			\nabla\left\{- Y^i\log(p)\right\}
			$
		}
	}
\end{algorithm}

	\paragraph{Training Process in AugDiff.}
To guide DAE to learn effective feature augmentation directions through the feature distribution of image augmentation, we first construct a training data set for the training process in AugDiff.
 According to the findings of \cite{tellez2019quantifying}, we choose six common and effective image augmentations: 1)~Random rotation, 2)~Random Elastic deformation, 3)~Random Affine transformation, 4)~Random Gaussian blurring, 5)~Random Color Jitter, 6)~Random Hematoxylin-Eosin-DAB (HED) Jitter. 
	
	In the training process of AugDiff,
 we employ the time-conditional UNet \cite{ronneberger2015u, Rombach_2022_CVPR} as DAE. 
Specifically, we randomly add a noise of level $t$ to the feature. Then the noisy feature is fed into DAE to predict the added noise. The objective of the optimization is to minimize the mean-squared error loss
 between the true noise and the predicted noise.
	The specific process is shown in Algorithm \ref{alg:train}.
\begin{algorithm}[ht]
		\small
		\caption{Training Process in AugDiff}
		\label{alg:train}
		\KwIn{
			The set of original feature $\mathcal{Z_O}$ and six sets of different augmented feature $\mathcal{Z_R}$, $\mathcal{Z_E}$, $\mathcal{Z_A}$, $\mathcal{Z_B}$, $\mathcal{Z_C}$, $\mathcal{Z_H}$.
		}
		\KwOut{Trained DAE $\epsilon_\theta$.}
		\While{not converged}{
			{\%1. Sample a data from the clean data distribution $q\left(\mathbf{z}_0\right)$ of $\mathcal{Z_O} \cup \mathcal{Z_R} \cup \mathcal{Z_E} \cup \mathcal{Z_A} \cup \mathcal{Z_B} \cup \mathcal{Z_C} \cup \mathcal{Z_H}$.}\\
			$\mathbf{z}_0 \sim q\left(\mathbf{z}_0\right)$ \\
			{\%2. Randomly sample noise from Gaussian distribution.}\\
$\boldsymbol{\epsilon} \sim \mathcal{N}(\mathbf{0}, \mathbf{I})$ \\
			{\%3. Randomly select a step and apply the diffusion process.}\\
			$t \sim \operatorname{Uniform}(\{1, \ldots, T\})$ \\
			$\mathbf{z}_t \sim \prod \limits_{i=1}^t q\left(\mathbf{z}_i|\mathbf{z}_{i-1}\right)$ \\
			{\%4. Take the gradient descent step.}\\
			$		\nabla_\theta\left\|\boldsymbol{\epsilon}-\boldsymbol{\epsilon}_\theta\left(\mathbf{z}_t, t\right)\right\|_2^2
			$
		}
\end{algorithm}
	
\paragraph{Sampling Process in AugDiff.}
Intuitively, the feature augmentation process should make no significant changes to the semantic information of input instance features. That is, the features of the tumor patch cannot be changed into the features of the normal patch after augmentation. 
	To accomplish this, we split the sampling process into two stages: 1)~$K$-step Diffusion and 2)~$K$-step Denoising.
	
To keep the original semantic information, the denoising process should start with a noise that contains the original semantic information instead of a randomly sampled noise. So, in the first stage, we apply the $K$-step Diffusion to the original features, where $K$ is less than $T$.
Then, in $K$-step Denoising, we employ the trained DAE to denoise the input features by $K$ steps. Since DAE is trained with a large number of augmented features, it can output the augmented version of the original features. Besides, it is worth mentioning that this two-stage sampling not only preserves the original semantic information but also significantly accelerates the sampling process.
	The sampling of AugDiff is shown in Algorithm~\ref{alg:sampling}.
\begin{algorithm}[ht]
	\small
	\caption{Sampling Process in AugDiff}
	\label{alg:sampling}
	\KwIn{
		The original bag ${Z_O}=\{\mathbf{z}^i_0\}^{N}_{i=1}$. Diffusion and Denoising step $K$ in the sampling process. 
	}
	\KwOut{Augmented bag $\widetilde{Z}$.}
	$\widetilde{Z} = \emptyset$ 
	
	\For{$\mathbf{z}_0^{i}$ in  $Z_O$}{
		{\%1. With $\mathbf{z}_0^{i}$ as the start, $K$ diffusion steps are taken.}\\
		$\mathbf{\widetilde{z}}_K^{i} \sim \prod \limits_{t=1}^K q\left(\mathbf{z}_t^{i}|\mathbf{z}_{t-1}^{i}\right)$ \\
		{\%2. Denoising diffusion process.}\\
		\For{$t=K,\ldots,1$}{
			{\% Use the trained $\epsilon_\theta$ and take the DDIM strategy.}\\
			$\mathbf{\widetilde{z}}_{t-1}^{i} \leftarrow \operatorname{DDIM}\left(\mathbf{\widetilde{z}}_{t}^{i}, t\right)$
		} 
		$\widetilde{Z}\leftarrow \widetilde{Z}\cup \{\mathbf{\widetilde{z}}_{0}^{i}\}$
	}
\end{algorithm}

\section{Experiments}
\begin{table}[]
	\small
	\begin{center}
	\begin{threeparttable}
	\setlength{\tabcolsep}{1.8mm}
	\begin{tabular}{c|cc|cccc}
		\toprule
		Datasets    & Size &  Num & Total & Train & Val & Test \\ \midrule
		SICAPv2 \cite{silva2020going}      & 512        & 121  & 155     & 95    & 27  & 33   \\
		UnitoPatho \cite{barbano2021unitopatho}    & 512        & 330  & 292     & 174   & 30  & 88   \\
		TMAs \cite{li2022differentiation}         & 224        & 115  & 786     & 472   & 118 & 196  \\ \bottomrule
	\end{tabular}

	\end{threeparttable}
	\end{center}
	\caption{\textbf{Dataset Description.} 
	We report data partitioning as well as statistics for the three used datasets.
	\textbf{Size.} 	
	The size of the patch utilized in the dataset, with each patch performing a series of data augmentations before feature extraction.
	\textbf{Num.} 
	The average number of patches contained in each WSI in the dataset.
	\textbf{SICAPv2.} SICAPv2 is a dataset containing prostate histological WSIs with Gleason-grade annotations. 
	We assign the Gleason grade labels to the following five categories: `0+0': 0, `3+3': 1, `3+4': 2, `4+3': 2, `4+4': 3, `3+5': 3, `5+3': 3, `4+5': 4, `5+4': 4, `5+5': 4.
	\textbf{UnitoPatho.} UnitoPatho is a dataset for the classification of colorectal cancer polyps that contains four categories: Normal tissue, Hyperplastic Polyp, Tubular Adenoma, and Tubulo-Villous Adenoma.
	\textbf{TMAs.} TMAs is a dataset for classifying pancreatic tissue with three categories: pancreatic ductal adenocarcinoma, chronic pancreatitis, and normal pancreatic tissue.
	}
	\label{tab:dataset}
\end{table}

\subsection{Implementation Details.}

\textit{(1)~Dataset Description.} We conducted experiments using three different cancer datasets, as shown in Table \ref{tab:dataset}, including SICAPv2 \cite{silva2020going} for prostate cancer, UnitoPatho \cite{barbano2021unitopatho} for colorectal cancer, and TMAs \cite{li2022differentiation} for pancreatic cancer. For UnitoPatho we used the official splitting, while for SICAPv2 and TMAs we used 4-fold cross-validation. 
\textit{(2)~MIL Methods.} We selected three classic MIL methods, including AMIL \cite{lu2021data}, LossAttn \cite{shi2020loss}, and DSMIL \cite{li2020dual}. 
The training is implemented using the open source code \cite{shao2021transmil}.
\textit{(3)~Feature Extraction.} We chose two different encoders for feature extraction, including ResNet18 \cite{He2015} and RegNetX\_004 \cite{radosavovic2020designing}. The pre-trained weights for the feature extractor are derived from the timm library \cite{rw2019timm}, and the features are 512- and 384-dimensional, respectively.
\textit{(4)~Comparison Experiments.} We compared with the no augmentation method, pseudo bag construction method used in \cite{zhang2022dtfd, shao2021weakly, bian2022multiple}, feature augmentation framework with Mixup \cite{mixup} which is used in \cite{yang2022remix, gadermayr2022mixup}, feature augmentation framework with GAN which is used in \cite{zaffar2022embedding}, and image-level augmentation methods. 
 \textit{(5) Details of AugDiff.} 
Our implementation is based on the \cite{Rombach_2022_CVPR}. The batch size of the training is 1200, the base learning rate is 5.0e-08, the sampling strategy adopts the DDIM \cite{song2020denoising}, the total steps $T$ is set to 20 or 30, and the number of sampling steps $K$ is set to 0.2$T$ or~0.4~$T$. 

\begin{table*}[]
	\small
	\begin{center}
        \scalebox{0.66}{
		\begin{tabular}{c|cccccccc|cccccccc}
			\toprule
			& \multicolumn{8}{c|}{SICAPv2 (ResNet18)}  & \multicolumn{8}{c}{SICAPv2 (RegNetX)}                                                                                                                                                                                      \\ \cline{2-17} 
			Augmentation       &
			\multicolumn{2}{c|}{\underline{AMIL \cite{lu2021data}}}                      & \multicolumn{2}{c|}{\underline{LossAttn \cite{shi2020loss}}}                  &  \multicolumn{2}{c|}{\underline{DSMIL \cite{li2020dual}}}                     & \multicolumn{2}{c|}{\underline{\textbf{Mean}}}&
			\multicolumn{2}{c|}{\underline{AMIL \cite{lu2021data}}}                      & \multicolumn{2}{c|}{\underline{LossAttn \cite{shi2020loss}}}                  &  \multicolumn{2}{c|}{\underline{DSMIL \cite{li2020dual}}}                     & \multicolumn{2}{c}{\underline{\textbf{Mean}}} \\
			& ACC         & \multicolumn{1}{c|}{AUC}         & ACC         & \multicolumn{1}{c|}{AUC}         & ACC         & \multicolumn{1}{c|}{AUC}               & ACC         &  \multicolumn{1}{c|}{AUC}  & ACC         & \multicolumn{1}{c|}{AUC}         & ACC         & \multicolumn{1}{c|}{AUC}         & ACC         & \multicolumn{1}{c|}{AUC}               & ACC         & AUC      \\ \midrule
			No Augmentation    & $0.292_{.11}$ & \multicolumn{1}{c|}{$0.684_{.12}$} & ${0.414}_{.09}$ & \multicolumn{1}{c|}{${0.725}_{.06}$} &  $0.341_{.08}$ & \multicolumn{1}{c|}{$0.696_{.02}$} & $0.349$       & $0.702$ & $0.340_{.06}$  & \multicolumn{1}{c|}{$0.719_{.03}$} & $0.382_{.06}$ & \multicolumn{1}{c|}{${0.722}_{.04}$} &  $\underline{0.414}_{.08}$ & \multicolumn{1}{c|}{$0.762_{.03}$} & $0.379$       & ${0.734}$       \\
			Patch Augmentation & $\underline{0.402}_{.08}$ & \multicolumn{1}{c|}{${0.750}_{.04}$}  & $0.382_{.05}$ & \multicolumn{1}{c|}{$0.714_{.07}$}  &  ${0.346}_{.03}$ & \multicolumn{1}{c|}{$0.714_{.03}$} & ${0.377}$       & ${0.726}$  & $0.401_{.06}$ & \multicolumn{1}{c|}{$0.723_{.04}$} & $\textbf{0.406}_{.04}$ & \multicolumn{1}{c|}{${0.716}_{.04}$} &  $0.400_{.07}$   & \multicolumn{1}{c|}{$0.761_{.04}$}  & $\underline{0.402}$       & $0.733$    \\ \midrule
			Pseudo Bag    & $\textbf{0.410}_{.08}$ & \multicolumn{1}{c|}{$0708_{.08}$} & ${0.398}_{.07}$ & \multicolumn{1}{c|}{${0.722}_{.06}$} &  $\textbf{0.410}_{.06}$ & \multicolumn{1}{c|}{$0.723_{.05}$} & $\textbf{0.406}$       & $0.718$ & $\underline{0.408}_{.04}$  & \multicolumn{1}{c|}{$0.731_{.04}$} & $0.370_{.09}$ & \multicolumn{1}{c|}{${0.715}_{.03}$} &  ${0.372}_{.01}$ & \multicolumn{1}{c|}{$0.755_{.04}$} & $0.383$       & ${0.734}$       \\
			Feature Mixup      & $0.278_{.10}$ & \multicolumn{1}{c|}{$0.676_{.11}$}& $0.285_{.12}$ & \multicolumn{1}{c|}{$0.648_{.14}$} &  $0.258_{.08}$ & \multicolumn{1}{c|}{$0.648_{.07}$} & $0.274$       & $0.657$  & ${0.405}_{.03}$ & \multicolumn{1}{c|}{$0.716_{.05}$} & $\underline{0.392}_{.14}$ & \multicolumn{1}{c|}{$0.673_{.10}$}  &  $\textbf{0.418}_{.09}$ & \multicolumn{1}{c|}{${0.764}_{.02}$} & $\textbf{0.405}$       & $0.718$   \\
			Pix2Pix       & $0.336_{.10}$ & \multicolumn{1}{c|}{$0.720_{.07}$}  & $0.349_{.05}$ & \multicolumn{1}{c|}{$0.714_{.06}$}  &  $0.332_{.02}$ & \multicolumn{1}{c|}{${0.725}_{.04}$} & $0.339$       &$ 0.720$   & ${0.404}_{.06}$ & \multicolumn{1}{c|}{${0.732}_{.06}$} & $0.323_{.05}$  & \multicolumn{1}{c|}{$0.666_{.05}$} &  $0.344_{.07}$ & \multicolumn{1}{c|}{$0.705_{.03}$} & $0.357$       & $0.701$   \\ \midrule
			AugDiff ($T$=20)           & ${0.372}_{.01}$ & \multicolumn{1}{c|}{$\underline{0.762}_{.05}$} & $\textbf{0.432}_{.13}$  & \multicolumn{1}{c|}{$\underline{0.733}_{.07}$} &  $\underline{0.382}_{.07}$ & \multicolumn{1}{c|}{$\textbf{0.752}_{.04}$} & $\underline{0.395}$       & $\underline{0.749}$  & $0.385_{.06}$ & \multicolumn{1}{c|}{$\textbf{0.745}_{.05}$} & ${0.345}_{.04}$ & \multicolumn{1}{c|}{$\underline{0.727}_{.03}$} &  $0.363_{.06}$  & \multicolumn{1}{c|}{$\underline{0.773}_{.03}$} & ${0.364}$       & $\underline{0.748}$     \\ 
			AugDiff ($T$=30)                       & ${0.393}_{.04}$ &\multicolumn{1}{c|}{$\textbf{0.765}_{.04}$} & $\underline{0.422}_{.12}$ &\multicolumn{1}{c|}{$\textbf{0.738}_{.08}$} & ${0.365}_{.06}$ &\multicolumn{1}{c|}{$\underline{0.749}_{.05}$} & 
			${0.393}$ &	$\textbf{0.751}$  & $\textbf{0.420}_{.05}$ & \multicolumn{1}{c|}{$\textbf{0.745}_{.04}$ }& $0.350_{.03}$ & \multicolumn{1}{c|}{$\textbf{0.737}_{.02}$}& $0.408_{.055}$ & \multicolumn{1}{c|}{$\textbf{0.786}_{.01}$ }& 
			${0.393}$&	$\textbf{0.756}$   \\                         
			\midrule

			& \multicolumn{8}{c|}{UnitoPatho (ResNet18)}     			& \multicolumn{8}{c}{UnitoPatho (RegNetX)}                                                                                                                                                                                      \\ \cline{2-17} 
			Augmentation       & 		 \multicolumn{2}{c|}{\underline{AMIL \cite{lu2021data}}}                      &  
			\multicolumn{2}{c|}{\underline{LossAttn \cite{shi2020loss}}}                  & 
			 \multicolumn{2}{c|}{\underline{DSMIL \cite{li2020dual}}}                     & \multicolumn{2}{c|}{\underline{\textbf{Mean}}}&
			\multicolumn{2}{c|}{\underline{AMIL \cite{lu2021data}}}                      &  
			\multicolumn{2}{c|}{\underline{LossAttn \cite{shi2020loss}}}                  & 
			 \multicolumn{2}{c|}{\underline{DSMIL \cite{li2020dual}}}                     & \multicolumn{2}{c}{\underline{\textbf{Mean}}} 
			 \\
			& ACC         & \multicolumn{1}{c|}{AUC}         & ACC         & \multicolumn{1}{c|}{AUC}         &  ACC         & \multicolumn{1}{c|}{AUC}         & ACC         & AUC        
			& ACC         & \multicolumn{1}{c|}{AUC}         & ACC         & \multicolumn{1}{c|}{AUC}         &  ACC         & \multicolumn{1}{c|}{AUC}         & ACC         & AUC     
			\\ \midrule
			No Augmentation    & $0.659$ & \multicolumn{1}{c|}{$0.895$} & $0.682$ & \multicolumn{1}{c|}{${0.904}$}  & $0.670$ & \multicolumn{1}{c|}{$0.828$} & $0.670$       & $0.876$     
			& $0.670$  & \multicolumn{1}{c|}{${0.858}$} & ${0.704}$ & \multicolumn{1}{c|}{$0.891$} &  $0.704$ & \multicolumn{1}{c|}{$0.868$} & $0.693$       & ${0.872}$ \\
   			Patch Augmentation & $\underline{0.727}$ & \multicolumn{1}{c|}{$0.889$}  & $0.704$ & \multicolumn{1}{c|}{${0.907}$}  & $0.670$ & \multicolumn{1}{c|}{${0.864}$} & ${0.700}$       & ${0.887}$    
			& $\underline{0.704}$ & \multicolumn{1}{c|}{$0.846$} & $0.693$ & \multicolumn{1}{c|}{${0.895}$} & $\underline{0.739}$   & \multicolumn{1}{c|}{${0.876}$}  & ${0.712}$       & ${0.872}$  \\ \midrule
			Pseudo Bag    & $0.716$ & \multicolumn{1}{c|}{$0.891$} & $\underline{0.716}$ & \multicolumn{1}{c|}{${0.906}$}  & $0.670$ & \multicolumn{1}{c|}{$0.827$} & $0.701$       & $0.875$     
			& $\underline{0.704}$  & \multicolumn{1}{c|}{${0.851}$} & ${0.704}$ & \multicolumn{1}{c|}{$0.891$} &  $0.727$ & \multicolumn{1}{c|}{$0.867$} & $0.712$       & ${0.870}$ \\
			Feature Mixup      & $0.682$ & \multicolumn{1}{c|}{${0.896}$}& ${0.704}$ & \multicolumn{1}{c|}{$0.900$} &  ${0.693}$ & \multicolumn{1}{c|}{$0.832$} & ${0.716}$       & ${0.888}$     
			& $0.670$ & \multicolumn{1}{c|}{$0.850$} & $\textbf{0.716}$ & \multicolumn{1}{c|}{$0.886$}  &  $0.693$ & \multicolumn{1}{c|}{$0.877$} & ${0.693}$       & $0.871$\\
			Pix2Pix       & $0.636$ & \multicolumn{1}{c|}{$0.854$}  & $0.648$ & \multicolumn{1}{c|}{$0.880$}  &  $0.682$ & \multicolumn{1}{c|}{$0.830$} & $0.665$       &$ 0.863$     
			& $0.693$ & \multicolumn{1}{c|}{$0.836$} & $0.625$  & \multicolumn{1}{c|}{$0.888$} &  $0.693$ & \multicolumn{1}{c|}{$0.860$} & $0.670$       & $0.861$\\ \midrule
			AugDiff ($T$=20)           & ${0.716}$ & \multicolumn{1}{c|}{$\textbf{0.906}$} & $\textbf{0.727}$  & \multicolumn{1}{c|}{$\textbf{0.915}$} & $\textbf{0.761}$ & \multicolumn{1}{c|}{$\underline{0.913}$} & $\underline{0.735}$       & $\textbf{0.911}$      
			& $\textbf{0.716}$ & \multicolumn{1}{c|}{$\textbf{0.899}$} & $\textbf{0.716}$ & \multicolumn{1}{c|}{$\underline{0.899}$} &  $\underline{0.739}$  & \multicolumn{1}{c|}{$\underline{0.900}$} & $\textbf{0.724}$       & $\underline{0.899}$\\ 
			AugDiff ($T$=30)           & $\textbf{0.761}$ & \multicolumn{1}{c|}{$\textbf{0.906}$} & $\underline{0.716}$  & \multicolumn{1}{c|}{$\underline{0.911}$} & $\textbf{0.761}$ & \multicolumn{1}{c|}{$\textbf{0.914}$} & $\textbf{0.746}$       & $\underline{0.910}$      
& $\underline{0.704}$ & \multicolumn{1}{c|}{$\underline{0.892}$} & $\textbf{0.716}$ & \multicolumn{1}{c|}{$\textbf{0.904}$} &  $\textbf{0.750}$  & \multicolumn{1}{c|}{$\textbf{0.905}$} & $\underline{0.723}$       & $\textbf{0.900}$\\                                       
\midrule

			& \multicolumn{8}{c|}{TMAs (ResNet18)} & \multicolumn{8}{c}{TMAs (RegNetX)}                                                                                                                                                                                       \\ \cline{2-17} 
			Augmentation       & 		 \multicolumn{2}{c|}{\underline{AMIL \cite{lu2021data}}}                      & \multicolumn{2}{c|}{\underline{LossAttn}\cite{shi2020loss}}                  &  \multicolumn{2}{c|}{\underline{DSMIL \cite{li2020dual}}}                     & \multicolumn{2}{c|}{\underline{\textbf{Mean}}} & 		 \multicolumn{2}{c|}{\underline{AMIL \cite{lu2021data}}}                      & \multicolumn{2}{c|}{\underline{LossAttn}\cite{shi2020loss}}                  &  \multicolumn{2}{c|}{\underline{DSMIL \cite{li2020dual}}}                     & \multicolumn{2}{c}{\underline{\textbf{Mean}}}\\
			& ACC         & \multicolumn{1}{c|}{AUC}         & ACC         & \multicolumn{1}{c|}{AUC}         & ACC         & \multicolumn{1}{c|}{AUC}         &  ACC         & AUC       			& ACC         & \multicolumn{1}{c|}{AUC}         & ACC         & \multicolumn{1}{c|}{AUC}         & ACC         & \multicolumn{1}{c|}{AUC}         &  ACC         & AUC         \\ \midrule
			No Augmentation    & ${0.821}_{.02}$ & \multicolumn{1}{c|}{$0.930_{.02}$} & ${0.822}_{.03}$ & \multicolumn{1}{c|}{$\underline{0.938}_{.02}$} &  $0.805_{.02}$ & \multicolumn{1}{c|}{$0.921_{.02}$} & $0.816$       & $0.930$  & $\textbf{0.833}_{.03}$  & \multicolumn{1}{c|}{$0.932_{.02}$} & $\underline{0.836}_{.02}$ & \multicolumn{1}{c|}{$\textbf{0.943}_{.02}$} &  $0.814_{.03}$ & \multicolumn{1}{c|}{$0.924_{.02}$} & ${0.828}$       & ${0.933}$     \\
   			Patch Augmentation & $0.814_{.03}$ & \multicolumn{1}{c|}{$0.933_{.02}$}  & $\underline{0.830}_{.02}$ & \multicolumn{1}{c|}{$0.937_{.01}$}  &  $\underline{0.824}_{.01}$ & \multicolumn{1}{c|}{${0.933}_{.01}$} & $\underline{0.823}$       & ${0.934}$   & $\underline{0.830}_{.03}$ & \multicolumn{1}{c|}{${0.935}_{.02}$} & $\textbf{0.842}_{.01}$ & \multicolumn{1}{c|}{$0.941_{.02}$} &  $\textbf{0.841}_{.03}$   & \multicolumn{1}{c|}{${0.933}_{.02}$}  & $\textbf{0.838}$       & ${0.936}$    \\ \midrule
			Pseudo Bag    & ${0.800}_{.03}$ & \multicolumn{1}{c|}{$0.927_{.03}$} & $\textbf{0.833}_{.03}$ & \multicolumn{1}{c|}{$\underline{0.938}_{.01}$} &  $0.804_{.02}$ & \multicolumn{1}{c|}{$0.926_{.01}$} & $0.812$       & $0.930$  & ${0.828}_{.03}$  & \multicolumn{1}{c|}{$0.926_{.02}$} & ${0.835}_{.02}$ & \multicolumn{1}{c|}{${0.934}_{.01}$} &  $0.819_{.01}$ & \multicolumn{1}{c|}{$0.924_{.01}$} & ${0.827}$       & ${0.928}$     \\
			Feature Mixup      & $0.812_{.03}$ & \multicolumn{1}{c|}{${0.934}_{.02}$}& $0.818_{.03}$ & \multicolumn{1}{c|}{$\underline{0.938}_{.02}$} &  $0.795_{.02}$ & \multicolumn{1}{c|}{$0.926_{.02}$} & $0.808$       & $0.933$   & $\underline{0.830}_{.03}$ & \multicolumn{1}{c|}{$0.933_{.02}$} & $0.828_{.02}$ & \multicolumn{1}{c|}{$\underline{0.942}_{.02}$}  &  $0.814_{.03}$ & \multicolumn{1}{c|}{$0.916_{.02}$} & ${0.824}$       & $0.930$  \\
			Pix2Pix       & $0.786_{.02}$ & \multicolumn{1}{c|}{$0.914_{.02}$}  & $0.809_{.03}$ & \multicolumn{1}{c|}{$0.930_{.02}$}  &  $0.742_{.03}$ & \multicolumn{1}{c|}{$0.892_{.02}$} & $0.779$       &$ 0.912$  & $0.780_{.04}$ & \multicolumn{1}{c|}{$0.910_{.03}$} & $0.793_{.05}$  & \multicolumn{1}{c|}{$0.916_{.02}$} &  $0.665_{.01}$ & \multicolumn{1}{c|}{$0.852_{.01}$} & $0.746$       & $0.893$    \\ \midrule
			AugDiff ($T$=20)           & $\underline{0.824}_{.03}$ & \multicolumn{1}{c|}{$\underline{0.939}_{.02}$} & $0.821_{.03}$  & \multicolumn{1}{c|}{$\textbf{0.939}_{.01}$} &  $\textbf{0.826}_{.01}$ & \multicolumn{1}{c|}{$\underline{0.935}_{.01}$} & $\textbf{0.824}$       & $\underline{0.938}$   & $0.821_{.01}$ & \multicolumn{1}{c|}{$\textbf{0.939}_{.01}$} & $0.821_{.01}$ & \multicolumn{1}{c|}{$0.938_{.01}$} &  $\underline{0.836}_{.01}$  & \multicolumn{1}{c|}{$\underline{0.935}_{.01}$} & $0.826$       & $\underline{0.937}$    \\
			AugDiff ($T$=30)           & $\textbf{0.831}_{.05}$ & \multicolumn{1}{c|}{$\textbf{0.944}_{.02}$} & $0.812_{.03}$  & \multicolumn{1}{c|}{${0.934}_{.01}$} &  ${0.819}_{.02}$ & \multicolumn{1}{c|}{$\textbf{0.941}_{.01}$} & ${0.821}$       & $\textbf{0.940}$  & $0.828_{.01}$ & \multicolumn{1}{c|}{$\underline{0.937}_{.01}$} & $\underline{0.836}_{.03}$ & \multicolumn{1}{c|}{$\underline{0.942}_{.02}$} &  ${0.823}_{.04}$  & \multicolumn{1}{c|}{$\textbf{0.940}_{.02}$} & $\underline{0.829}$       & $\textbf{0.940}$     \\
			\bottomrule
		\end{tabular} }
	\end{center}
	\caption{\textbf{Results of SICAPv2, UnitoPatho and TMAs.} We compare five popular MIL training frameworks based on three widely used MIL models, two different feature extractors, and three different cancer datasets. Metrics include micro ACC and macro AUC, where \textbf{bold} represents the highest value, and \underline{underlined} represents the second-highest value. All results in SICAPv2 and TMAs were obtained through 4-fold cross-validation and described in the form of $\text{Mean}_{\text{std}}$. Besides, all results reported in UnitoPatho are based on the official splitting and evaluated over the official test dataset. 
	}
	\label{tab:sicap}
\end{table*}

\subsection{Main Results}
\begin{figure}[h]
	\centering
	\includegraphics[width=0.95\linewidth]{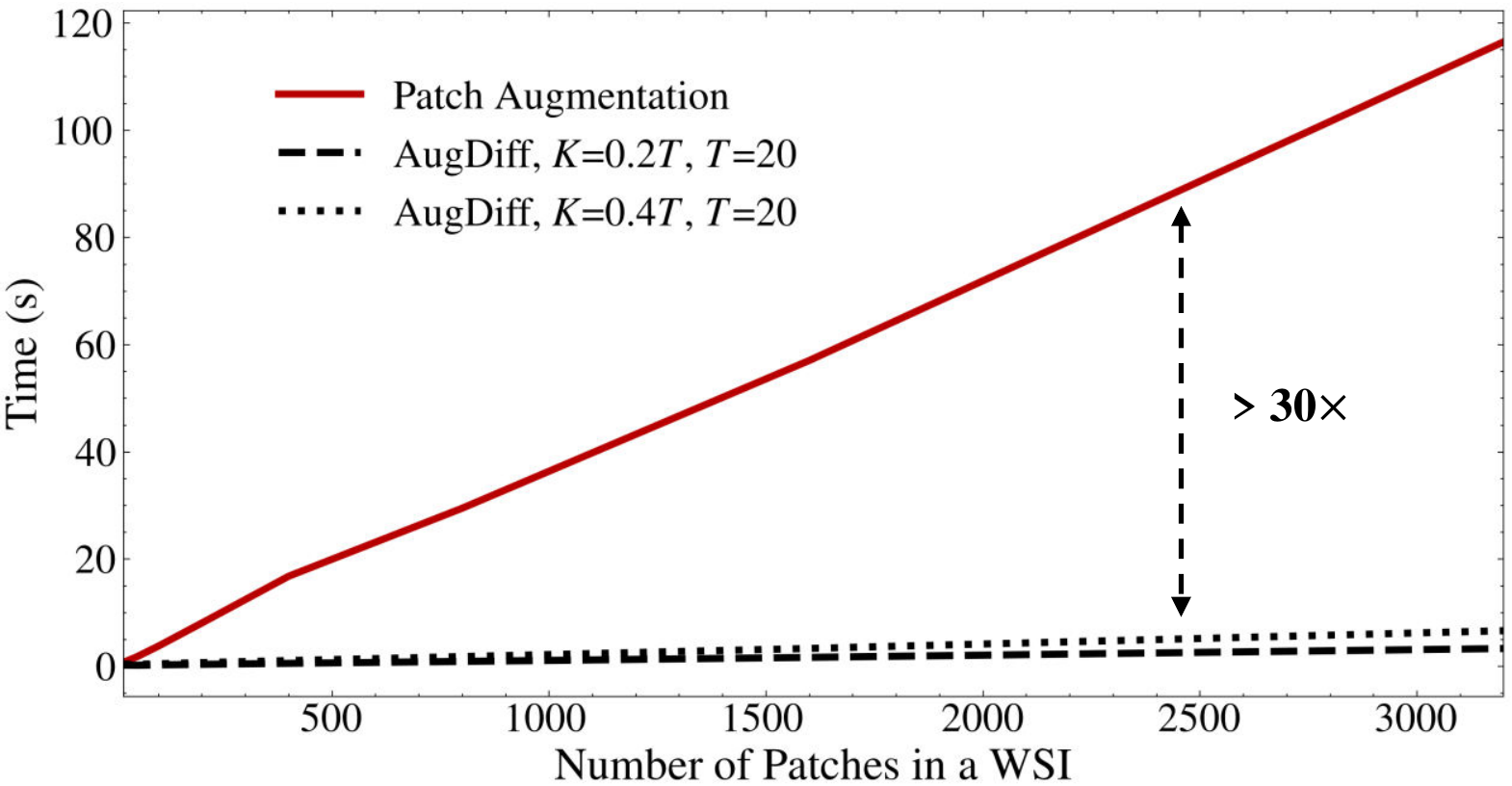}
	\caption{\textbf{Speed comparison of Patch Augmentation and AugDiff in MIL training.} 
	We compare the speed of augmenting a WSI with various patch numbers under different augmentation frameworks. The vertical ordinate denotes the time needed for the Patch Augmentation framework or AugDiff, and the horizontal coordinate denotes the total number of patches in a WSI. See the appendix for specific experimental details.
	}
	\label{fig:speed}
\end{figure}

To demonstrate the general applicability of our framework, we conducted experiments using three prevalent MIL methods on three cancer datasets. The results are shown in Table \ref{tab:sicap}.
To ensure a fair comparison, we did not augment the test sets during the evaluation of all frameworks.
To demonstrate the superiority of the proposed framework in terms of speed over Patch Augmentation, we performed speed tests and summarized the results in Figure~\ref{fig:speed}. We can obtain the following observations.

1) Compared to MIL training without data augmentation (No Augmentation), current feature augmentation methods failed to consistently improve model performance to the same extent as Patch Augmentation (\eg, UnitoPatho (RegNetX), mean AUC, Feature Mixup -0.1\%, Pseudo Bag -0.2\%, Pix2Pix -1.1\%). 
The linear mixing operation of Mixup can easily lead to unrealistic features. The pseudo-bag method makes it easy to introduce label noise while expanding the training data. In GAN feature generation training, weak discriminators tend to result in weak generators and consequently unrealistic features. In contrast, Patch Augmentation is an image-level augmentation method that produces more realistic images and extracts more realistic augmented features.

2) Though Patch Augmentation consistently facilitates the MIL training, AugDiff is more effective in terms of both performance and speed (\eg, SICAPv2 (ResNet18), mean AUC, $T$=20 +2.3\%. UnitoPatho (RegNetX), mean AUC,  $T$=20 +1.5\%). 
For performance, the redundancy of pixel space limits Patch Augmentation, leading to limited changes in feature space after Patch Augmentation. In contrast, the AugDiff-learned feature augmentation is more informative and representative. 
For speed, AugDiff performs more than 30 times faster than the Patch Augmentation framework (as shown in Figure~\ref{fig:speed}). The time-consuming feature extraction process required for the Patch Augmentation framework is unnecessary in our AugDiff framework because it performs augmentation at the feature level.  

3) Benefiting from the step-by-step generation of the Diffusion Model, AugDiff can be further improved by increasing the number of sampling steps. Specifically, the parameter $T$ determines the total number of model sampling steps and their fineness. A larger $T$ can often help AugDiff generate better features (\eg, SICAPv2 (RegNetX), mean AUC, compared to $T$=20, $T$=30 +0.8\%). A smaller $T$ can also help AugDiff achieve a balance between performance and time while achieving superior results than competing methods and ensuring a faster sampling rate. 

\subsection{Ablation Study}
The MIL training is significantly influenced by $K$, as it determines how much the original semantic information is retained during the sampling process. To investigate the effect of parameter $K$ under various scenarios, we tested four values of $K$ under two datasets, two feature extractors, and three MIL models. All results are presented in Table~\ref{tab:ablation}. 

A larger $K$ tends to fully exert the augmentation effect of AugDiff. Compared to $K=0.1T$, the performance of AugDiff with a larger $K$, such as $0.2T/0.3T/0.4T$, could be significantly improved (\eg, SICAPv2 (ResNet18), compared to $0.1T$, $0.2T$ +0.8\%, $0.3T$ +1.4\%, $0.4T$ +1.6\%). 
Besides, the different choices of feature extractors and datasets could slightly affect the optimal choice of parameter $K$.  For example, in the SICAPv2 dataset, when ResNet18 is used to extract features, the mean AUC will gradually increase with the increase of $K$. In contrast, when RegNetX is used to extract features, the optimal mean AUC is achieved near the $K$ value of $0.2T$ or $0.3T$. In the UnitoPatho dataset, using different feature extractors, the performance of the MIL model typically improves as $K$ increases, with $K = 0.4T$ often providing the best average performance.

In the appendix, we also discussed the impact of the conditional-guided mechanism on AugDiff. We compared the performance of AugDiff with unconditional-guided and conditional-guided and visualized the different types of augmentation under the conditional-guided mechanism. In summary, we found that conditional sampling has no significant effect on feature augmentation. 

\begin{table}[]
	\small
\begin{center}
\scalebox{0.66}{
	\setlength{\tabcolsep}{1.4mm}
	\begin{tabular}{c|cccc|cccc}
		\toprule
		\multirow{2}{*}{Augmentation} & \multicolumn{4}{c|}{\underline{SICAPv2 (ResNet18)}}         &  \multicolumn{4}{c}{\underline{UnitoPatho (ResNet18)}} \\
		& AMIL        & LossAttn     & DSMIL        & \textbf{Mean}  &  AMIL     & LossAttn    & DSMIL    & \textbf{Mean}    \\ \midrule
		$K=0.1T$                        & $0.752_{.05}$ & ${0.729}_{.07}$ & $0.719_{.05}$ & 
		$0.733$                       &$0.895$    & $0.902$      & $0.887$   & $0.895$   \\
		$K=0.2T$                        & ${0.765}_{.05}$ & $0.727_{.07}$ & $0.732_{.04}$ & 
		$0.741$                         &$0.912$    & $0.905$      & $0.913$   & $0.910$   \\
		$K=0.3T$                        & ${0.764}_{.05}$ & $0.726_{.07}$ & ${0.750}_{.05}$ & 
		${0.747}$                         &$0.908$    & $0.900$      & $0.918$   & ${0.909}$   \\
		$K=0.4T$                        & $0.762_{.05}$ & ${0.733}_{.07}$ & ${0.752}_{.04}$ & 
		$\textbf{0.749}$                        &$0.906$    & $0.915$      & $0.913$   & $\textbf{0.911}$   \\ 
\midrule
		\multirow{2}{*}{Augmentation} & \multicolumn{4}{c|}{\underline{SICAPv2 (RegNetX)}}           & \multicolumn{4}{c}{\underline{UnitoPatho (RegNetX)}}  \\
		& AMIL         & LossAttn     & DSMIL       & \textbf{Mean}  &AMIL     & LossAttn    & DSMIL    & \textbf{Mean}    \\ \midrule
		$K=0.1T$                        & $0.727_{.04}$ & $0.728_{.03}$ & $0.762_{.02}$ & 
		$0.739$                       &$0.859 $   & $0.890$      & $0.872$   & $0.874$   \\
		$K=0.2T$                        & $0.745_{.05}$ & $0.727_{.03}$ & $0.773_{.03}$ & 
		$\textbf{0.748}$                         &$0.873 $   & $0.880$      & $0.907$   & $0.887$   \\
		$K=0.3T$                        & $0.754_{.03}$  & $0.727_{.04}$ & $0.762_{.02}$ & 
		$\textbf{0.748}$                         &$0.880$    & $0.894$      & $0.900$   & $0.891$  \\
		$K=0.4T$                        & $0.750_{.03}$ & $0.734_{.04}$ & $0.747_{.03}$  & 
		$0.744$                        &$0.899$    & $0.899$      & $0.900$   & $\textbf{0.899}$   \\         \bottomrule
	\end{tabular} }
\end{center}
	\caption{\textbf{Effects of different settings in AugDiff.} We discuss the parameter $K$ in the AugDiff sampling process over the SICAPv2 and UnitoPatho datasets. For SICAPv2, the reported \textbf{AUC} are described in the form of $\text{Mean}_{\text{std}}$. For UnitoPatho, the \textbf{AUC} are reported over the official test dataset.
}
\label{tab:ablation}
\end{table}

\subsection{Visualization}

We utilized the UMAP\cite{2018arXivUMAP} technique to perform dimensionality reduction on the embedding space of ResNet18 and then compared the augmented feature distribution of Patch Augmentation with that of AugDiff in the reduced embedding space. 
Details of our implementation and more visualization results  are provided in the appendix. 
Figure~\ref{fig:visualization} intuitively demonstrate the rationality of AugDiff and its superiority over Patch Augmentation. 
First, as shown in (a) and (b), as the augmentation rounds of AugDiff increase, the distribution of AugDiff's augmented features (blue dots) approaches that of Patch Augmentation's augmented features (red dots), indicating that AugDiff could effectively simulate image-level augmentation in the embedding space.
Second, as shown in (b) and supplementary Figure~\ref{fig:visualization_trend}, the blue dots form a more distinct structure than the red dots under the same augmentation rounds, indicating that the augmentation quality of AugDiff is better than Patch Augmentation.
Furthermore, due to its better efficiency, AugDiff could generate more augmented samples than Patch Augmentation at the same time cost, thus improving the robustness and generalization of the MIL model. This is illustrated in (c), where the blue dots almost entirely cover the red dots.

\begin{figure}
  \centering
  \begin{subfigure}[b]{0.32\linewidth}
   \centering
   \includegraphics[width=\linewidth]{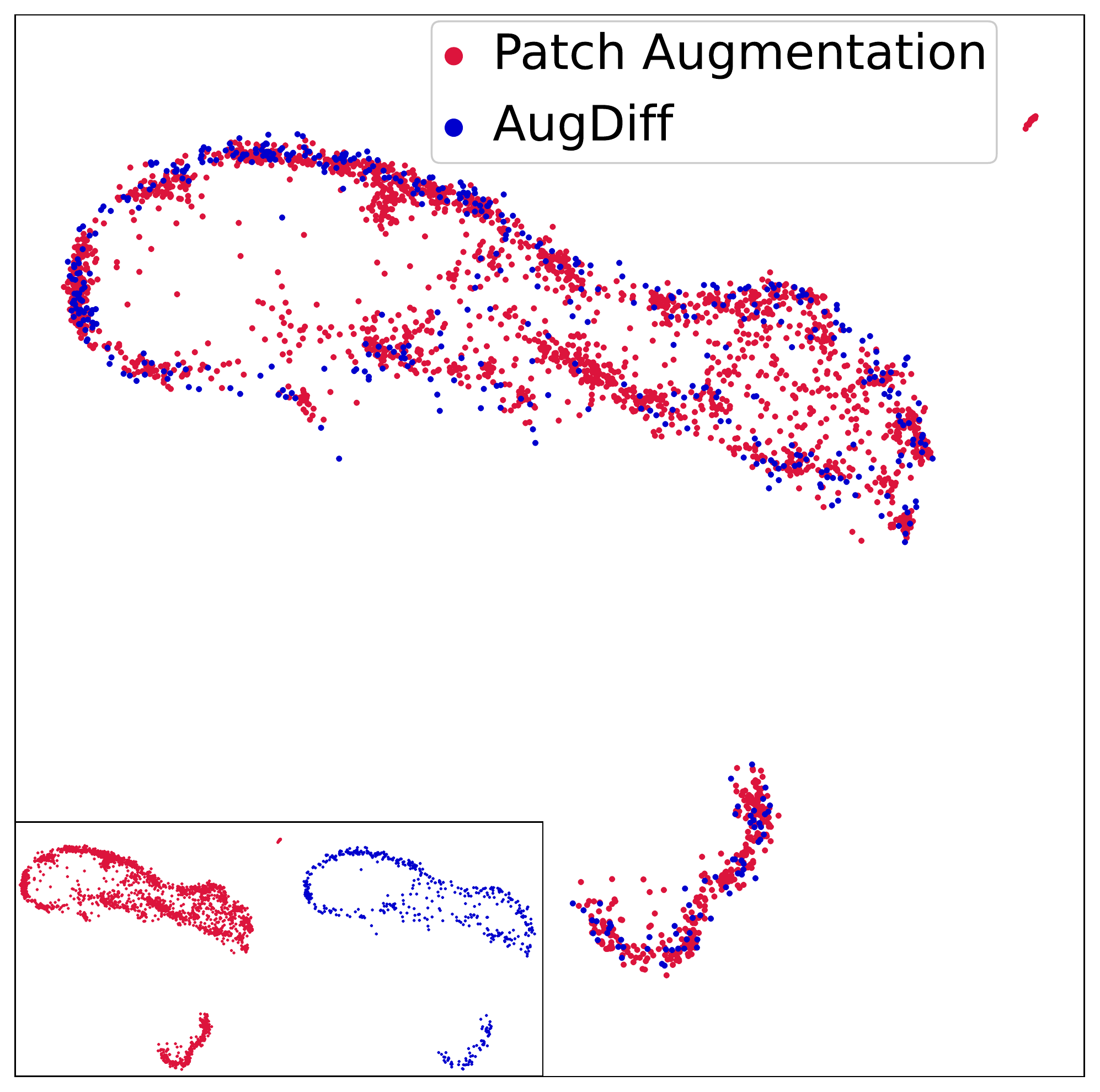}
   \caption{}
   \label{fig:visual_5epochs}
  \end{subfigure}
  \begin{subfigure}[b]{0.32\linewidth}
   \centering
   \includegraphics[width=\linewidth]{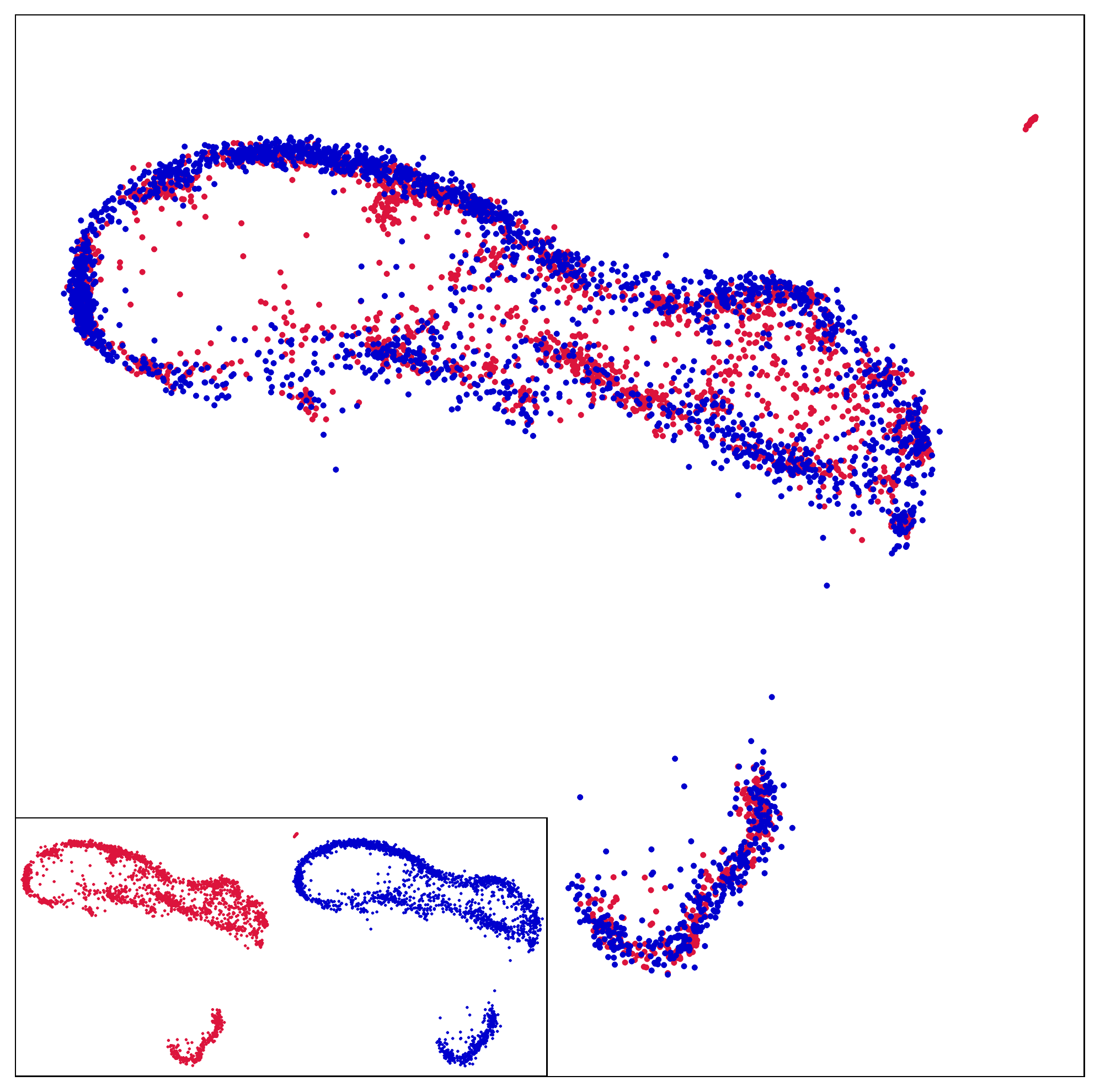}
   \caption{}
   \label{fig:visual_25epochs}
  \end{subfigure}
  \begin{subfigure}[b]{0.32\linewidth}
   \centering
   \includegraphics[width=\linewidth]{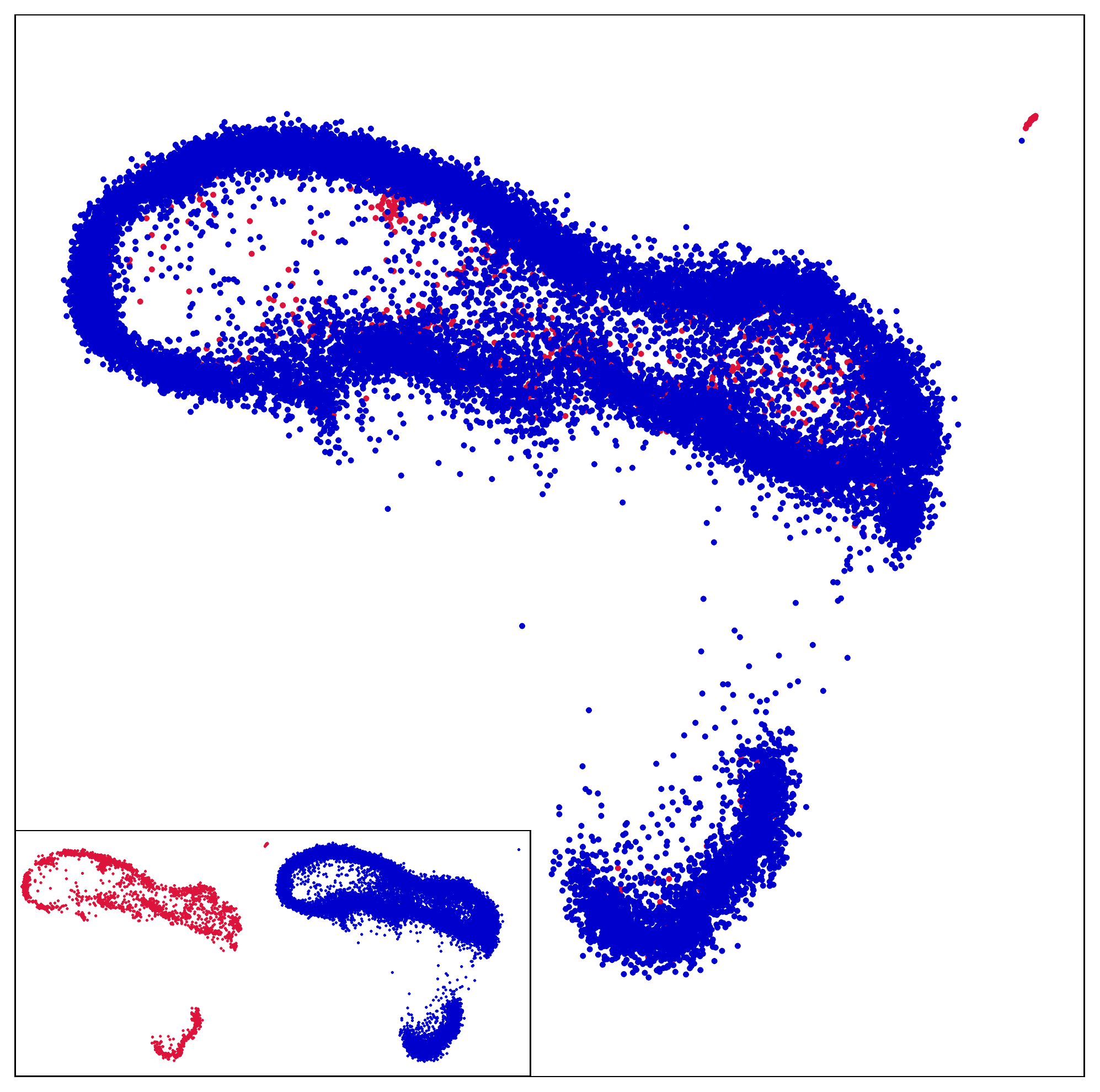}
   \caption{}
   \label{fig:visual_50epochs}
  \end{subfigure}
  \caption{
   \textbf{Comparison between AugDiff and Patch Augmentation in the 
   ResNet18 embedding space.}
   \textbf{(a)}~AugDiff (1$\times$) vs. Patch Augmentation (5$\times$).
   \textbf{(b)}~AugDiff (5$\times$) vs. Patch Augmentation (5$\times$).
   \textbf{(c)}~AugDiff (50$\times$) vs. Patch Augmentation (5$\times$).
   The red and blue dots represent features of Patch Augmentation and AugDiff, respectively. $n\times$ denotes augmentation rounds.
  }
  \label{fig:visualization}
 \end{figure}

\begin{table}[h]
	\small
	\begin{center}
 \scalebox{0.66}{
    \setlength{\tabcolsep}{1.1mm}
    \begin{tabular}{c|cccc|cccc}
        \toprule
        \multirow{2}{*}{Augmentation} & \multicolumn{4}{c|}{\underline{SICAPv2 (ResNet18)}}          & \multicolumn{4}{c}{\underline{SICAPv2 (RegNetX)}} \\
        & AMIL        & LossAttn    & DSMIL      & \textbf{Mean}  & AMIL   & LossAttn    & DSMIL    & \textbf{Mean}    \\ \midrule
        PA (1$\times$)                       & ${0.750}_{.04}$  & $0.714_{.07}$  &  $0.714_{.03}$     & ${0.726}$ & $0.723_{.04}$ & ${0.716}_{.04}$ &  $0.761_{.04}$        & $0.733$    \\
        PA (5$\times$)                         & $0.758_{.04}$ & ${0.732}_{.07}$ & $0.720_{.02}$ & 
        $0.737$                      &${0.728}_{.02}$    & ${0.729}_{.04}$      & ${0.758}_{.04}$   & $0.738$   \\
        PA (10$\times$)                      & $0.758_{.04}$ & ${0.725}_{.06}$ & ${0.726}_{.03}$ & 
        ${0.736}$                        &$0.737_{.02}$ & ${0.732}_{.04}$ & ${0.755}_{.05}$ & 
        ${0.741}$  \\  
        AugDiff           & $0.762_{.05}$ & ${0.733}_{.07}$ & ${0.752}_{.04}$ & 
        $\textbf{0.749}$ &               ${0.745}_{.05}$ & ${0.727}_{.03}$ &  ${0.773}_{.03}$   & $\textbf{0.748}$     \\ \bottomrule
    \end{tabular} 
  }
	\end{center}
	\caption{\textbf{Performance comparison of multiple rounds of Patch Augmentation vs. AugDiff.} In the SICAP dataset, we conduct additional rounds of patch augmentation to test the performance. PA denotes Patch Augmentation. $n\times$ denotes the number of times each patch augmentation method is applied to each patch. The \textbf{AUC} is reported in the form of $\text{Mean}_{\text{std}}$.
	}
	\label{tab:discussion_patchaug}
\end{table}

\section{Discussion}

\paragraph{Superiority of AugDiff over More Patch Augmentation.} 
To demonstrate the significant advantage of AugDiff in generating high-quality augmented features, in addition to its speed, we conducted a performance comparison between multiple rounds of patch augmentation and AugDiff.
We tested the performance of the MIL model under 1, 5, and 10 rounds of image augmentation on the SICAPv2 dataset. The more rounds of patch augmentation, the more features are utilized for MIL training with the Patch Augmentation framework. It should be noted that each round of patch augmentation includes six common types of patch augmentation. Furthermore, since the MIL training process adopts the early stopping mechanism, 10 rounds of patch augmentation can result in a similar amount of data expansion as AugDiff. The specific results are shown in Table \ref{tab:discussion_patchaug}. 

Compared with one round of patch augmentation, as the augmentation rounds increase, the performance of the MIL model improves (\eg, SICAPv2 (RegNetX), compared to PA (1$\times$), PA (5$\times$) +0.5\%, PA (10$\times$) +0.8\%). However, we also found that the improvement based on the number of image augmentation rounds is often limited (\eg, SICAPv2 (RegNet18), compared to PA (1$\times$), PA (5$\times$) +1.1\%, PA (10$\times$) +1.0\%). In addition, by comparing 10 rounds of patch augmentation with AugDiff, we discovered that the feature augmentation strategy based on AugDiff can obtain more competitive results over various feature extractors and MIL models, demonstrating the high quality of AugDiff augmented features.

\begin{table}[h]
	\small
	\begin{center}
 \scalebox{0.66}{
		\setlength{\tabcolsep}{1.1mm}
		\begin{tabular}{c|cccc|cccc}
			\toprule
			\multirow{2}{*}{Augmentation} & \multicolumn{4}{c|}{\underline{SICAPv2 (ResNet18)}}          & \multicolumn{4}{c}{\underline{UnitoPatho (ResNet18)}} \\
			& AMIL        & LossAttn    & DSMIL      & \textbf{Mean}  & AMIL   & LossAttn    & DSMIL    & \textbf{Mean}    \\ \midrule
			No Augmentation                        & $0.684_{.12}$ & ${0.725}_{.06}$ & $0.696_{.02}$ & 
$0.702$                     &$0.895$    & $0.904$      & $0.828$   & $0.876$   \\
			SimSiam                        & $0.606_{.10}$ & ${0.622}_{.07}$ & $0.615_{.08}$ & 
			$0.614$                      &$0.856$    & $0.883$      & $0.902$   & $0.880$   \\
			AugDiff                      & $0.762_{.05}$ & ${0.733}_{.07}$ & ${0.752}_{.04}$ & 
			$\textbf{0.749}$                        &$0.906$    & $0.915$      & $0.913$   & $\textbf{0.911}$   \\  \bottomrule
		\end{tabular} }
	\end{center}
	\caption{\textbf{Discussion on self-supervised learning and feature augmentation.} We choose SimSiam \cite{Chen_2021_CVPR} in self-supervised training as a comparison. For SICAPv2, the reported \textbf{AUC} are described in the form of $\text{Mean}_{\text{std}}$. For UnitoPatho, the \textbf{AUC} are reported over the official test dataset.
	}
	\label{tab:discussion_simsiam}
\end{table}

\begin{table}[h]
	\small
	\begin{center}
 \scalebox{0.66}{
		\setlength{\tabcolsep}{1.1mm}
		\begin{tabular}{c|cccc|cccc}
			\toprule
			\multirow{2}{*}{Augmentation} & \multicolumn{4}{c|}{\underline{SICAPv2 (ResNet18)}}          & \multicolumn{4}{c}{\underline{UnitoPatho (ResNet18)}} \\
			& AMIL        & LossAttn     & DSMIL      & \textbf{Mean}  &  AMIL     & LossAttn    & DSMIL   & \textbf{Mean}    \\ \midrule
			Patch Augmentation                        & $0.750_{.04}$ & ${0.714}_{.07}$ & $0.714_{.03}$ & 
			$0.726$                      &$0.889$    & $0.907$      & $0.864$   & $0.887$   \\
   			SICAPv2                      & $0.762_{.05}$ & ${0.733}_{.07}$ & ${0.752}_{.04}$ & 
			$\textbf{0.749}$                        &$0.907$    & $0.908$      & $0.915$   & ${0.910}$   \\
			UnitoPatho                        & $0.759_{.05}$ & ${0.731}_{.08}$ & $0.726_{.06}$ & 
			$0.739$                         &$0.906$    & $0.915$      & $0.913$   & $\textbf{0.911}$   \\ \midrule
			\multirow{2}{*}{Augmentation} & \multicolumn{4}{c|}{\underline{SICAPv2 (RegNetX)}}          & \multicolumn{4}{c}{\underline{UnitoPatho (RegNetX)}}  \\
			& AMIL        & LossAttn    & DSMIL      & \textbf{Mean}  &AMIL     & LossAttn   & DSMIL    & \textbf{Mean}    \\ \midrule
			Patch Augmentation                        & $0.723_{.04}$ & ${0.716}_{.04}$ & $0.761_{.04}$ & 
$0.733$                       &$0.846$    & $0.895$      & $0.876$   & $0.872$   \\
			SICAPv2                        & $0.745_{.05}$ & $0.727_{.03}$ & $0.773_{.03}$  & 
			$\textbf{0.748}$                        &$0.906$    & $0.890$      & $0.906$   & $\textbf{0.901}$   \\
			UnitoPatho                        & $0.746_{.05}$ & $0.726_{.04}$ & $0.772_{.04}$ & 
			$\textbf{0.748}$                        &$0.899$    & $0.899$      & $0.900$   & $0.899$  \\ \bottomrule
		\end{tabular} }
	\end{center}
	\caption{\textbf{Discussion on generalization ability of AugDiff.} We discuss about AugDiff's ability to generalize to external datasets after being pretrained on specific datasets. In the \textit{Augmentation} column, \textit{UnitoPatho} and \textit{SICAPv2} stand for pre-trained data sets used by AugDiff. In SICAPv2, the reported \textbf{AUC} are described in the form of $\text{Mean}_{\text{std}}$. In UnitoPatho, the \textbf{AUC} are reported in the official test dataset.}
	\label{tab:discussion_alter}
\end{table}

\paragraph{Superiority of AugDiff over Self-supervised Learning Framework.} 
AugDiff, as a training framework guided by image augmentation, is more promising than Self-Supervised Learning (SSL) methods guided by agent tasks. Specifically, SSL methods are typically based on preset agent tasks, require large-scale datasets, and have a more disordered training procedure. In contrast, AugDiff is guided by the mapping of image augmentation in feature space, which is a more direct and effective guide. 
To compare the applicability of the two frameworks, we selected SimSiam \cite{Chen_2021_CVPR}, a commonly used SSL method, as a comparison. Specifically, in the downstream dataset, we first pre-trained ResNet18 based on SimSiam and then introduced the pre-trained feature extractor into the traditional MIL framework (No Augmentation). The corresponding results are shown in Table \ref{tab:discussion_simsiam}.

Since the SSL training process is more disordered, it has a more urgent need for the size of the dataset. 
In the relatively small dataset SICAPv2, SSL has a less pre-training effect than UnitoPatho. In the relatively large dataset UnitoPatho, the SSL barely exceeds the traditional MIL framework. 
More importantly, the combination of SSL and MIL is often used in the feature extractor. SSL needs to at least get better results than large-scale ImageNet pre-trained feature extractors. The combination of AugDiff and MIL is used to do the feature augmentation. It only needs to expand the diversity of the features extracted by the ImageNet pre-trained encoder to improve the MIL, which is a more straightforward and less data-demanding task. As a result, with stronger prior information for training and greater suitability for MIL, the AugDiff framework can have a better performance than SSL over datasets of different sizes.

\paragraph{Superiority of AugDiff in Generalization Ability.} 
We conducted cross-tests over two different cancer datasets to verify the generalization ability of pre-trained AugDiff on external datasets. The experimental setup is as follows: 
1) AugDiff pre-trained on UnitoPatho is used for feature augmentation over the SICAPv2 dataset; 2) AugDiff pre-trained on SICAPv2 is used for feature augmentation over the UnitoPatho dataset. 
 Due to the fact that AugDiff's training and testing rely on fixed feature dimensions, there are no cross-tests between feature extractors of different dimensions (\eg, ResNet18 and RegNetX).
Table \ref{tab:discussion_alter} summarizes all experimental results. 
We found that the pre-trained AugDiff has good feature augmentation capabilities when tested on external datasets, and all the external test results are higher than Patch Augmentation. Besides, since the SICAPv2 dataset may have a more diverse patch morphology, AugDiff was better pre-trained on the SICAPv2 dataset.
Note that the morphology often varies between different cancer datasets \cite{ciga2022self}. Since AugDiff's training is guided by image augmentation, it avoids overfitting to the specific dataset. Thus, the pre-trained AugDiff can be deployed to a wider variety of datasets and applications.

\section{Conclusion}
In this work, we present a novel feature augmentation framework based on Diffusion Model (DM), AugDiff, for the MIL training process in the WSI-related tasks.
Restricted by limited WSI-level training data, image augmentation is often used to assist MIL training. However, tens of thousands of patches in WSIs limit image augmentation, and it can only assist MIL in an offline manner, which is an inefficient and sub-optimal solution. Feature augmentation is a promising solution with online augmentation, but unrealistic and unstable generations always limit performance.
To solve the above issues, we introduce the diversity-generating ability of DM into the MIL training, where the input features can be augmented in each training epoch. Besides, the step-by-step generation characteristic of DM can control the retention of semantic information during the augmentation sampling process.
We perform sufficient tests over three cancer datasets, two different feature extractors, and three widely used MIL models. The comparison results show the proposed AugDiff can significantly improve the current framework, and the ablation study shows the rationality of the sampling process. Furthermore, we highlight the high quality of augmented features and the superiority of AugDiff over the SSL method. 
Discussions on generalization and analysis of visualizations further demonstrate that AugDiff has excellent prospects.
In the future, we will evaluate the performance of AugDiff over larger datasets and employ faster sampling techniques to improve the speed further.

{\small
	\bibliographystyle{ieee_fullname}
	\bibliography{egbib}
}

\clearpage

\section{Appendix}

\begin{figure}[h]
	\centering
	\includegraphics[width=0.95\linewidth]{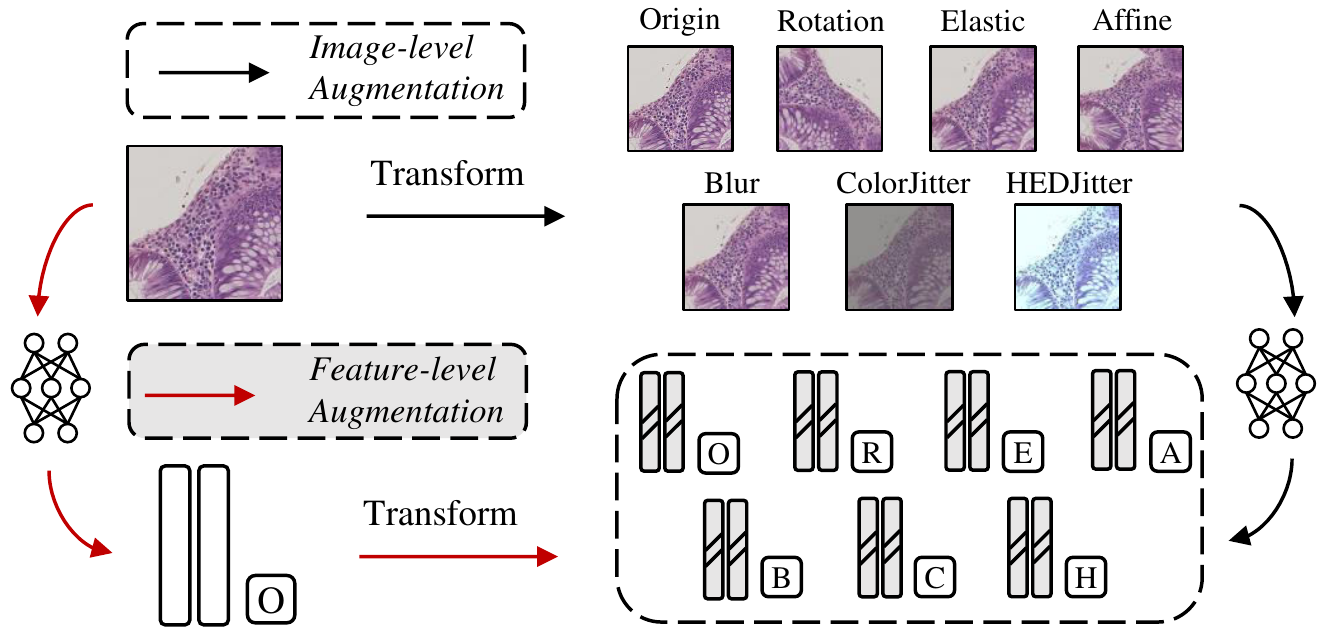}
	\caption{\textbf{Construct the feature training Dataset for Denoising Autoencoder with Patch Augmentation and Feature Extraction.} 
	We show how to construct a training dataset for feature augmentation and give an example of different image augmentations.
	}
	\label{fig:imageaug}
\end{figure}

\subsection{Supplementary Implementation Details}
\paragraph{Image Augmentation.} The following parameters are chosen for image augmentation: 1)~Random rotation, which includes rotation by 90 degrees and vertical and horizontal mirroring. 2)~Random Elastic deformation with alpha equal to 2 and sigma equal to 0.06. 3)~Random Affine transformation with an alpha value of 0.1. 4)~Random Gaussian blurring with a radius of 0.5 to 1.5. 5)~Random Color Jitter includes brightness and contrast image perturbation with a brightness intensity ratio between 0.65 and 1.35 and a contrast intensity ratio between 0.5 and 1.5. 6)~Random Hematoxylin-Eosin-DAB (HED) Jitter with intensity ratios is 0.05. In Figure \ref{fig:imageaug}, we show an example of data augmentation.

\paragraph{Speed Comparison.} We compare the speed between the Patch Augmentation framework and AugDiff. Three steps are involved in the Patch Augmentation framework: 1) reading every patch in WSI, 2) making the patch augmentation, and 3) extracting every patch's features. The following steps are part of AugDiff: 1) reading the features of every patch in WSI and 2) performing a feature data augmentation. We test the speed of AugDiff with different settings.

\subsection{Effects of Condition-guided Mechanism }

The ablation study discuss the effect of the conditional-guided mechanism in AugDiff. The conditional guidance mechanism refers to direct guidance of feature augmentation during the training and sampling, which is widely used in image generation. In the test, we evaluate three MIL methods of two types of feature extractors on two datasets. All results are presented in Table~\ref{tab:ablation_condition}.

Although the conditional guidance mechanism is widely used in image generation, it does not perform significantly better than unconditional guidance in the task of feature augmentation. Table~\ref{tab:ablation_condition} shows that conditional and unconditional augmentation perform similarly on different datasets and feature extractors. 
There are the following reasons: 1)~The core of feature augmentation to facilitate MIL model training is generating additional features to enlarge the training data set. The impact of increasing the number of features is greater than that of conditional and unconditional augmentation. In addition, the diverse generation capabilities of the Diffusion model further weaken the difference between conditional and unconditional augmentation.  2)~Compared with the difference between different augmentation types in image augmentation, the difference between distinct feature augmentation may be more challenging to learn. The existing simple conditional mechanism may still need to improve. Besides, the visual analysis in the appendix further validates our results.

\begin{table}[h]
	\small
\begin{center}
\scalebox{0.66}{
	\setlength{\tabcolsep}{1.4mm}
	\begin{tabular}{c|cccc|cccc}
		\toprule
		\multirow{2}{*}{Augmentation} & \multicolumn{4}{c|}{\underline{SICAPv2 (ResNet18)}}         &  \multicolumn{4}{c}{\underline{UnitoPatho (ResNet18)}} \\
		& AMIL        & LossAttn     & DSMIL        & \textbf{Mean}  &  AMIL     & LossAttn    & DSMIL    & \textbf{Mean}    \\ \midrule
		Uncondition                        & $0.762_{.04}$ & ${0.728}_{.07}$ & ${0.745}_{.04}$ & 
${0.745}$                        &$0.908$    & $0.916$      & $0.913$   & $\textbf{0.912}$   \\ 
		Condition                        & $0.762_{.05}$ & ${0.733}_{.07}$ & ${0.752}_{.04}$ & 
$\textbf{0.749}$                        &$0.906$    & $0.915$      & $0.913$   & ${0.911}$   \\
\midrule
		\multirow{2}{*}{Augmentation} & \multicolumn{4}{c|}{\underline{SICAPv2 (RegNetX)}}           & \multicolumn{4}{c}{\underline{UnitoPatho (RegNetX)}}  \\
		& AMIL         & LossAttn     & DSMIL       & \textbf{Mean}  &AMIL     & LossAttn    & DSMIL    & \textbf{Mean}    \\ \midrule
		Uncondition                       & $0.741_{.05}$ & $0.722_{.04}$ & $0.775_{.04}$ & 
		${0.746}$                         &$0.904$    & $0.898$      & $0.903$   & $\textbf{0.902}$   \\
Condition                        & $0.745_{.05}$ & $0.727_{.03}$ & $0.773_{.03}$ & 
$\textbf{0.748}$  &$0.899$    & $0.899$      & $0.900$   & ${0.899}$           \\             \bottomrule
	\end{tabular} }
\end{center}
	\caption{\textbf{Effects of different settings in AugDiff.} We discuss the effects of the condition-guided mechanism in the AugDiff sampling process over the SICAPv2 and UnitoPatho datasets. In SICAPv2, the reported \textbf{AUC} are described in the form of $\text{Mean}_{\text{std}}$. In UnitoPatho, the \textbf{AUC} are reported in the official test dataset.
}
\label{tab:ablation_condition}
\end{table}

\subsection{UMAP-based Visualization}
\paragraph{Implementation of UMAP.} 
We utilize the open-source implementation \cite{mcinnes2018umap-software} of UMAP with the following hyperparameters: neighbors=50, dist=0.1, and random\_state=42. We select a WSI (17B0024162) from the SICAPv2 dataset containing 88 patches. We apply each of the six Patch Augmentation to all patches for 5 rounds, resulting in a total of $5 \times 6 \times 88=2,640$ image-level augmented patches (corresponding to the red dots in Figure \ref{fig:visualization}a, \ref{fig:visualization}b and \ref{fig:visualization}c). Additionally, we apply AugDiff with each of the six conditions (excluding condition=0) to all patches for 100 times, resulting in a total of $100 \times 6 \times 88=52,800$ feature-level augmented patches. It should be noted that we used $T=30$ and $K=0.4T$ in the sampling process in AugDiff. Then, we take 1, 5, and 50 patches from feature-level augmented patches under each condition for visualization, respectively (corresponding to the blue dots in Figure \ref{fig:visualization}a, \ref{fig:visualization}b and \ref{fig:visualization}c, respectively).
\paragraph{More Visualization Results.}

\begin{figure*}
    \centering
  \begin{subfigure}[b]{0.32\linewidth}
       \centering
       \includegraphics[width=\linewidth]{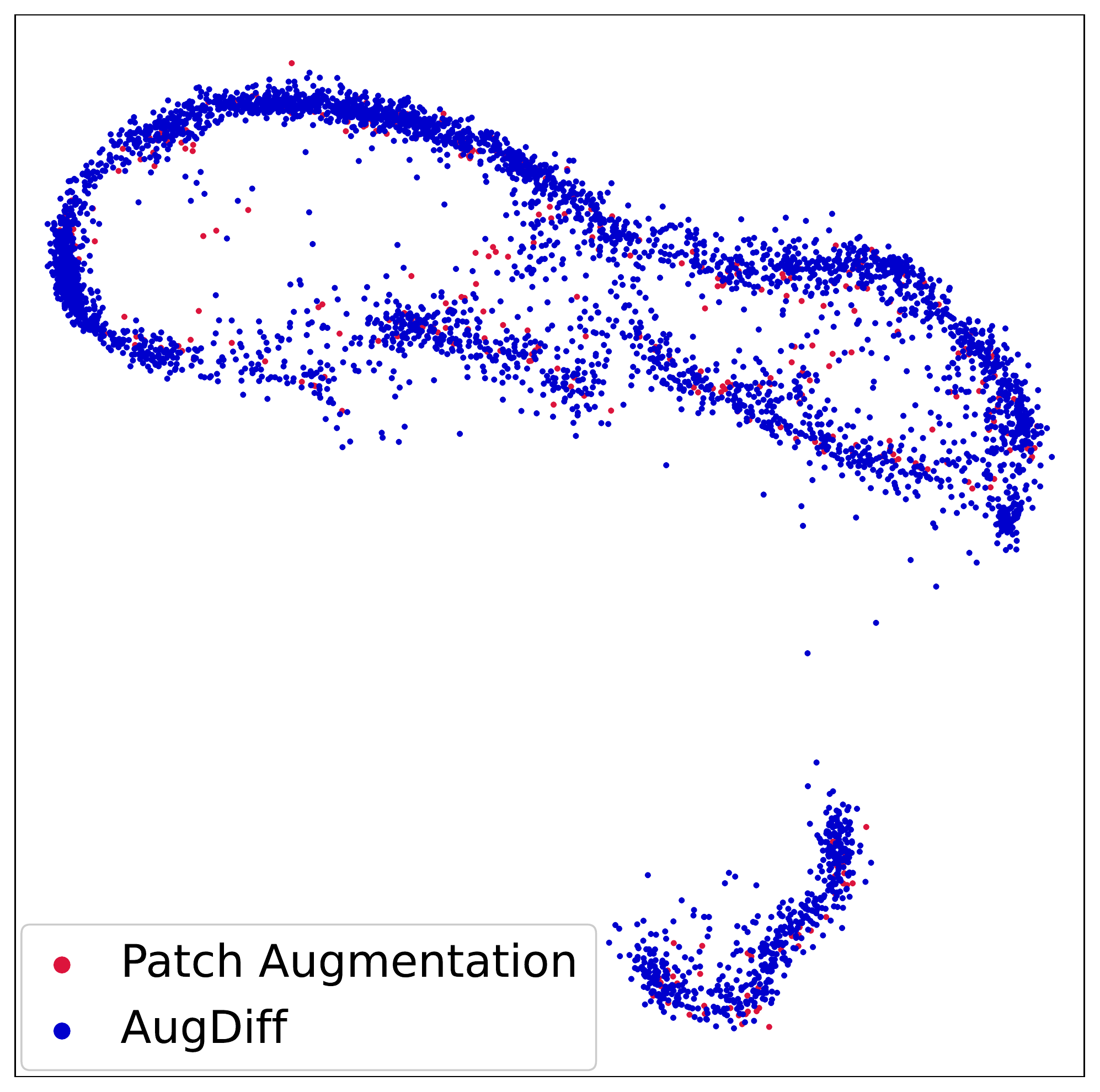}
       \caption{}
  \end{subfigure}
  \begin{subfigure}[b]{0.32\linewidth}
       \centering
       \includegraphics[width=\linewidth]{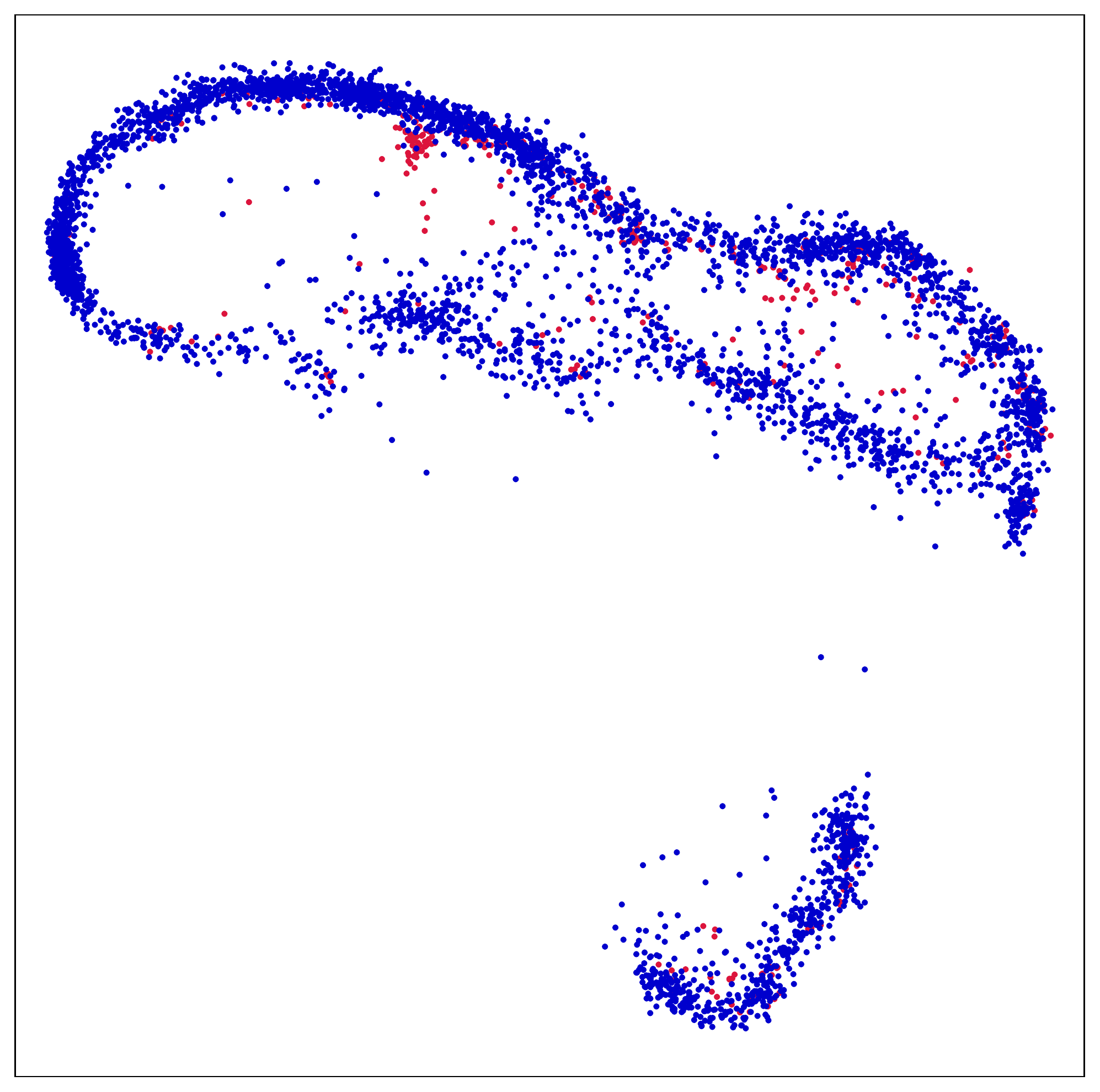}
       \caption{}
  \end{subfigure}
  \begin{subfigure}[b]{0.32\linewidth}
       \centering
       \includegraphics[width=\linewidth]{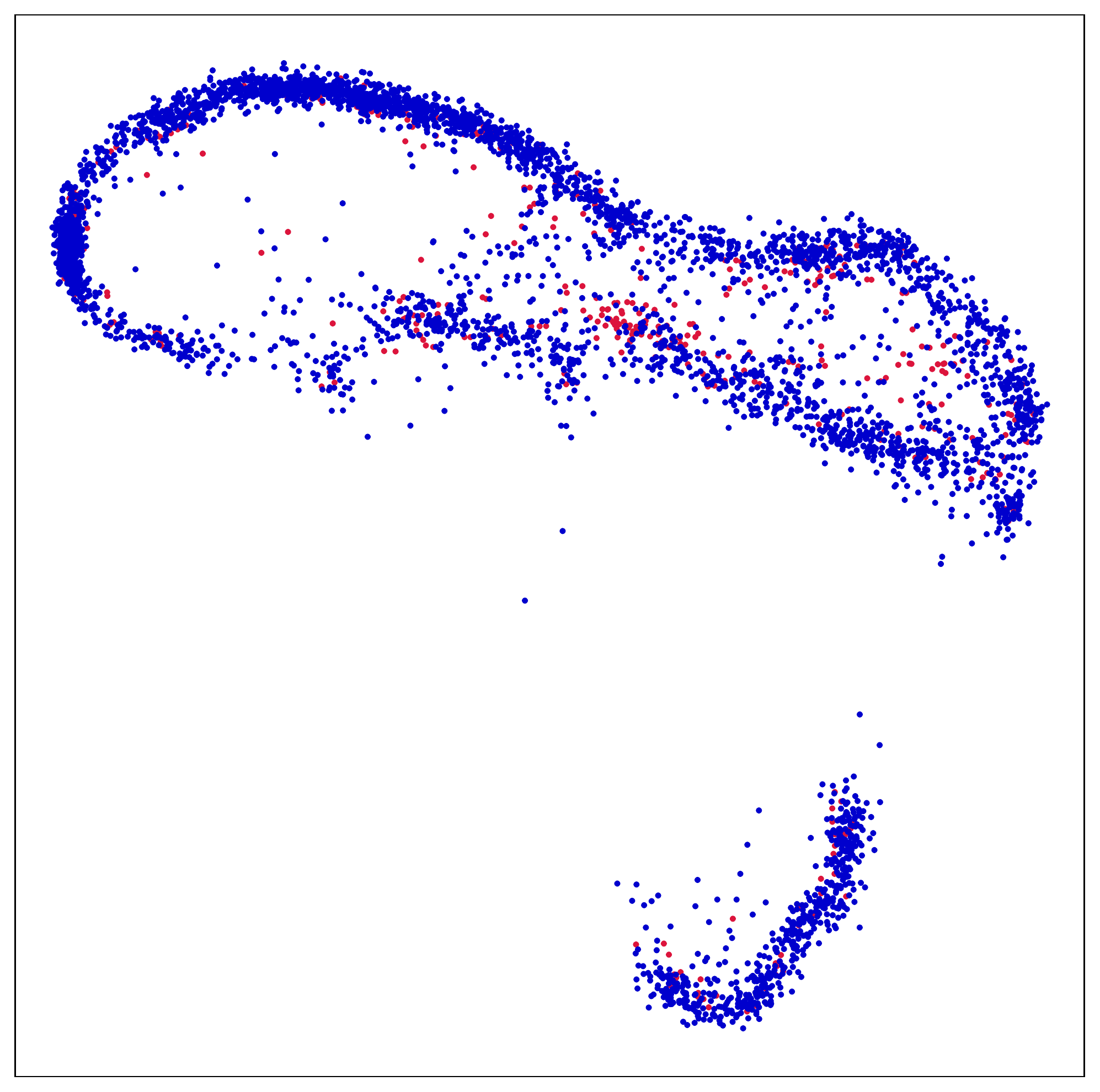}
       \caption{}
  \end{subfigure}
  \begin{subfigure}[b]{0.32\linewidth}
       \centering
       \includegraphics[width=\linewidth]{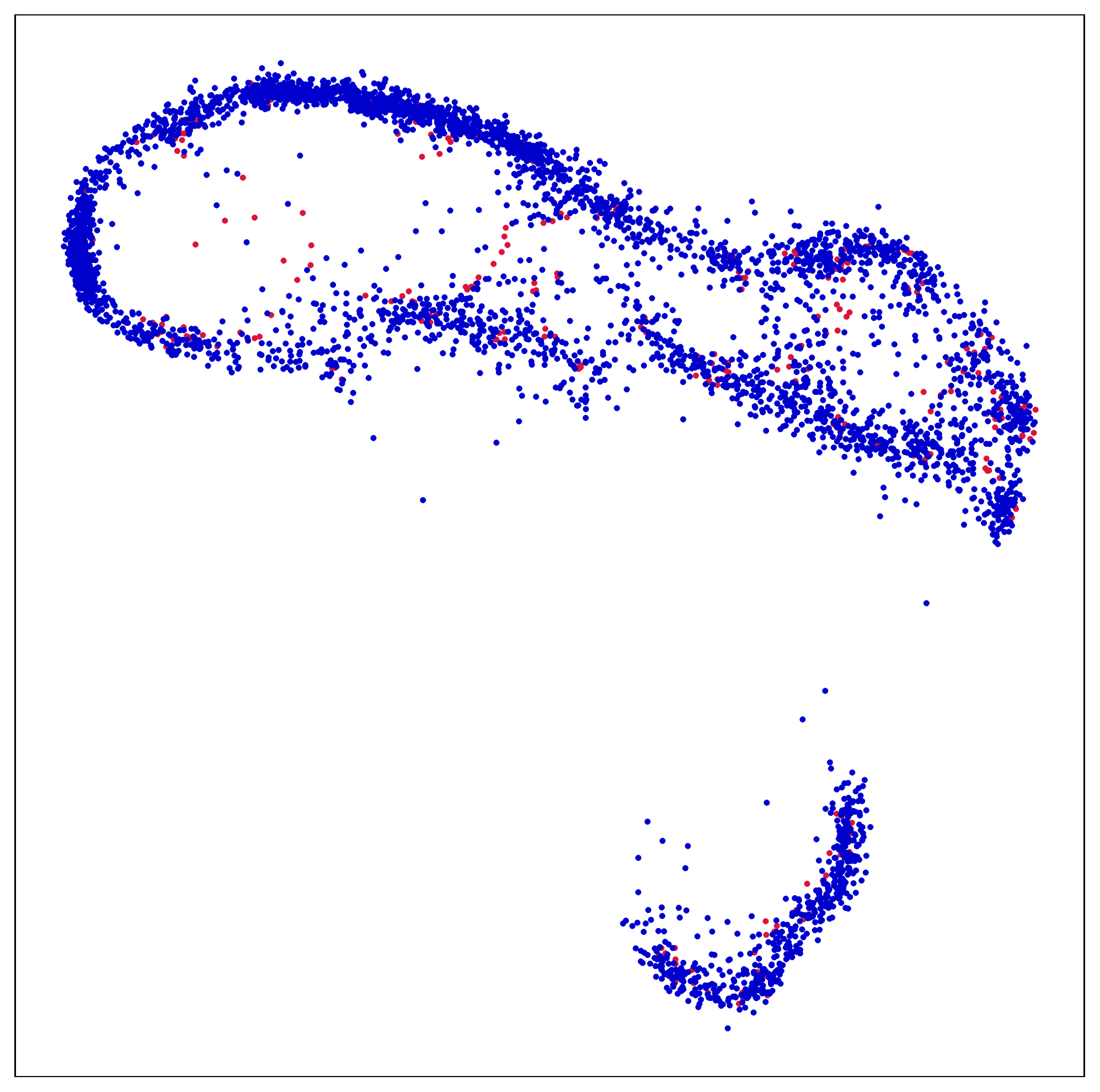}
       \caption{}
  \end{subfigure}
  \begin{subfigure}[b]{0.32\linewidth}
       \centering
       \includegraphics[width=\linewidth]{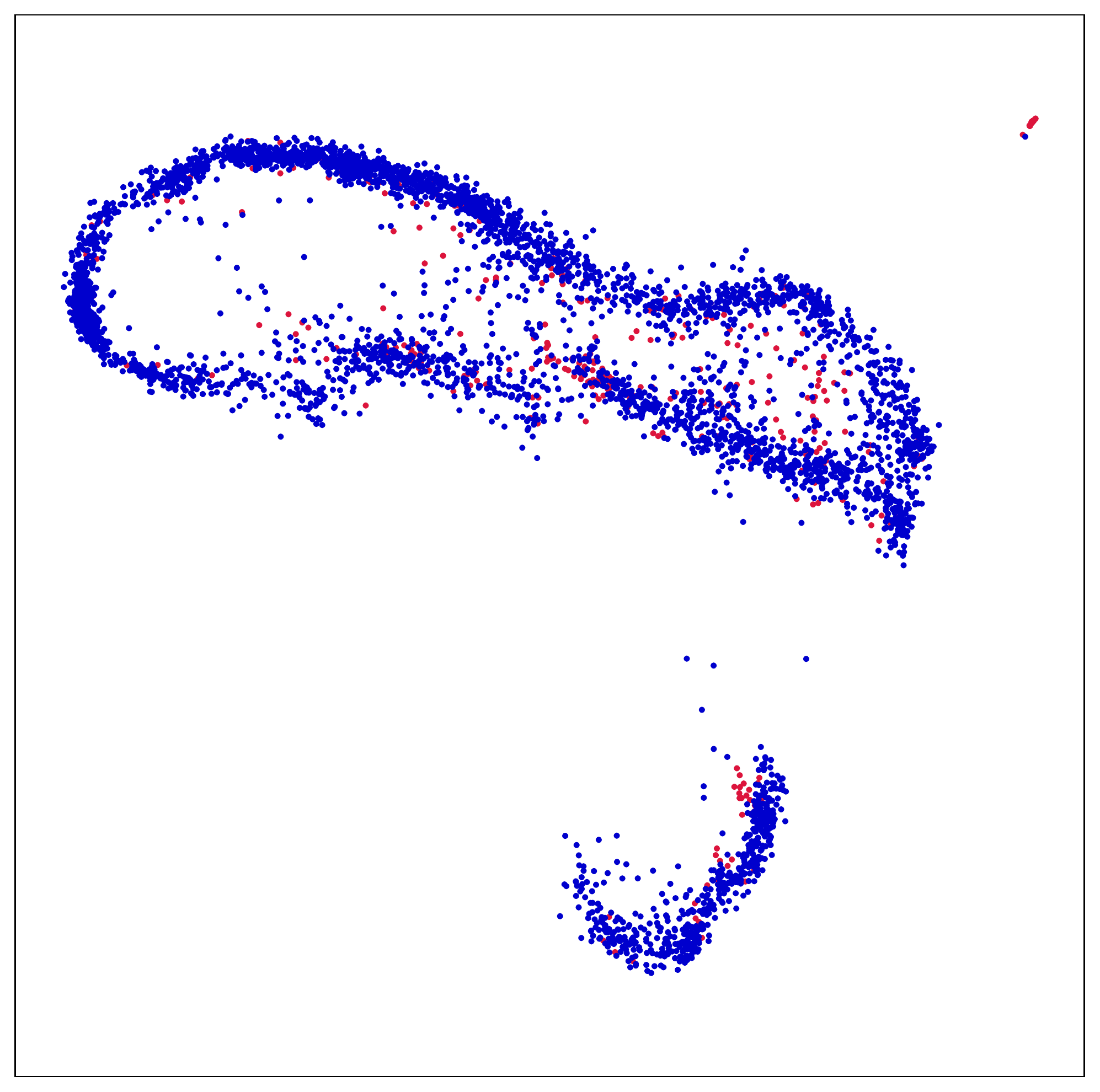}
       \caption{}
  \end{subfigure}
  \begin{subfigure}[b]{0.32\linewidth}
       \centering
       \includegraphics[width=\linewidth]{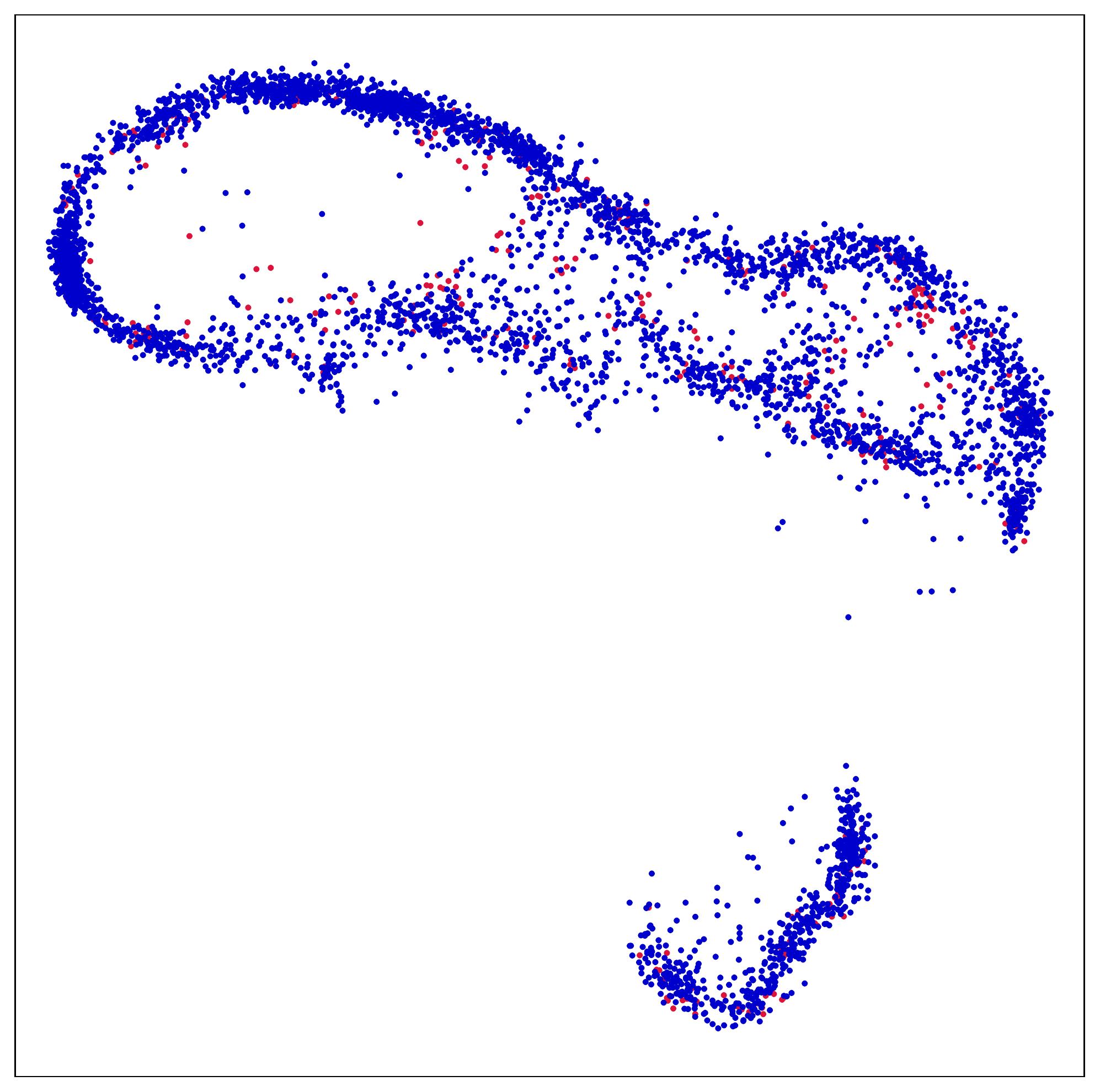}
       \caption{}
  \end{subfigure}
  
  \caption{
   \textbf{Comparison between AugDiff and Patch Augmentation in the low-dimensional ResNet18 embedding space when adopting distinct augmentation types.}
    \textbf{(a)} AugDiff of condition=1 vs. Patch Augmentation of Rotation. 
    \textbf{(b)} AugDiff of condition=2 vs. Patch Augmentation of Elastic. 
    \textbf{(c)} AugDiff of condition=3 vs. Patch Augmentation of Affine. 
    \textbf{(d)} AugDiff of condition=4 vs. Patch Augmentation of Blur. 
    \textbf{(e)} AugDiff of condition=5 vs. Patch Augmentation of ColorJitter. 
    \textbf{(f)} AugDiff of condition=6 vs. Patch Augmentation of HEDJitter. 
   The red dots represent augmented samples of Patch Augmentation (5$\times$), and the blue dots represent augmented samples of AugDiff (50$\times$). $n\times$ denotes augmentation rounds.
  }
  \label{fig:visualization_condition_1to6}
 \end{figure*}

\begin{figure*}[h]
	\centering
	\includegraphics[width=0.95\linewidth]{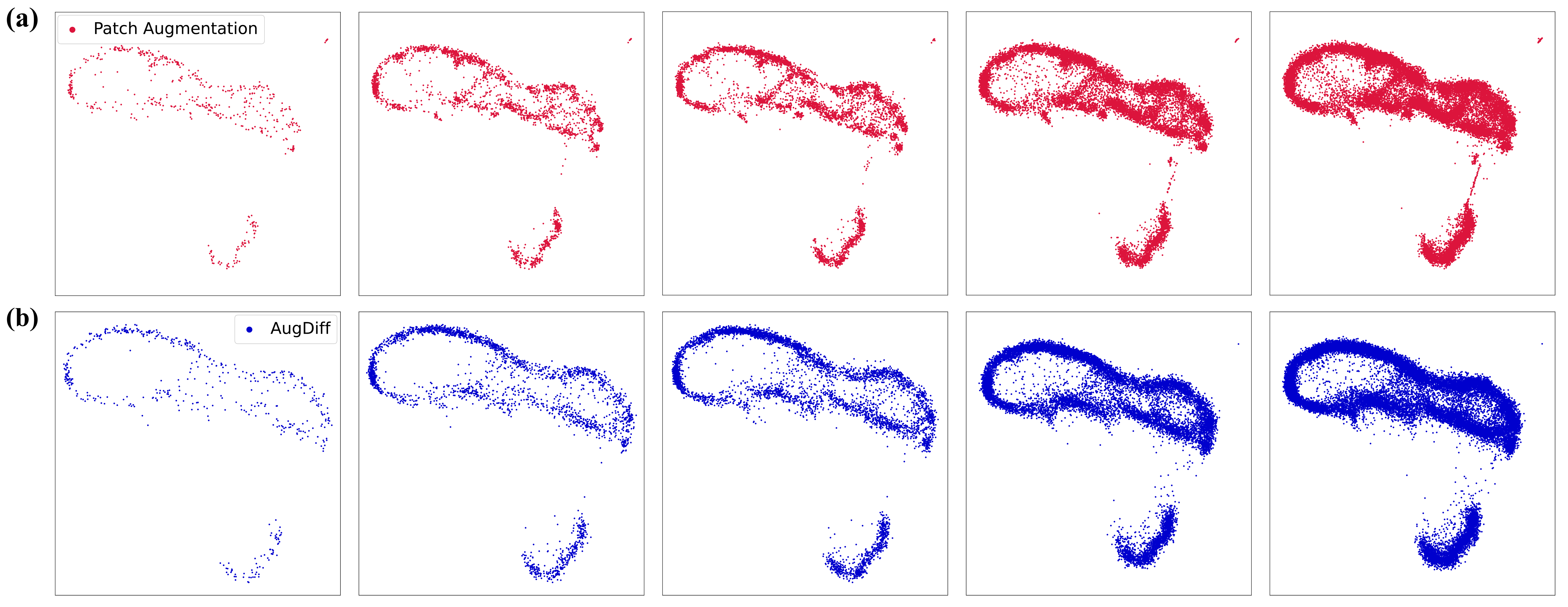}
	\caption{\textbf{The distribution trends of Patch Augmentation and AugDiff.} 
        \textbf{(a)} We apply Patch Augmentation for 1, 5, 10, 25, 50 rounds and separately visualized the feature distribution of generated instances, respectively. 
        \textbf{(b)} We apply AugDiff for 1, 5, 10, 25, 50 rounds and separately visualized the feature distribution of generated instances, respectively. 
        }
	\label{fig:visualization_trend}
\end{figure*}

To better understand AugDiff, we additionally use UMAP for two visualizations: 1) the augmented feature distributions of AugDiff under different augmentation types, \ie, under conditions from 1 to 6 (Figure \ref{fig:visualization_condition_1to6}), and 
2) the augmented feature distributions of AugDiff and Patch Augmentation under different augmentation rounds.
For 1), we apply AugDiff under each of six conditions for 50 rounds, then separately visualize the 4,400 generated features and the 440 image-level augmented features of the corresponding augmentation type. It can be observed from Figure \ref{fig:visualization_condition_1to6} that different types of image-level augmentation lead to remarkably similar distributions (see the red dots in a-f). Meanwhile, AugDiff can effectively simulate each image-level augmentation under the corresponding condition, which makes the augmented features distribution of Augdiff under various conditions close to each other (see the blue dots in a-f). That explains why the condition-guides mechanism has little impact on the model performance, as shown in Table \ref{tab:ablation_condition}.
For 2), we applied AugDiff and Patch Augmentation 1, 5, 10, 25, and 50 rounds, respectively. We separately visualized the augmented AugDiff and Patch Augmentation samples under each augmentation time. As shown in Figure \ref{fig:visualization_trend}, the augmented features' density distribution trends of AugDiff and Patch Augmentation are slightly different. Specifically, as the augmentation rounds increase, AugDiff puts higher density onto the border of the manifold, while Patch Augmentation puts a higher density inside the manifold. According to the results in  Table~\ref{tab:discussion_patchaug}, we hypothesize that the augmented samples distributed on the manifold border are of better quality because the performance gain of Patch Augmentation decreases with the increment of augmentation rounds. Therefore, AugDiff can bring more significant performance gain with less time cost than Patch Augmentation as the augmentation rounds increase.
\end{document}